%% file: main.tex
\documentclass[10pt,twocolumn,letterpaper]{article}

\usepackage{cvpr}              
\usepackage{indentfirst}
\usepackage{mathrsfs}

\input{preamble}

\usepackage[pagebackref,breaklinks,colorlinks]{hyperref}
\usepackage{multirow}
\usepackage{adjustbox}
\usepackage{amsmath}
\usepackage{pythonhighlight}
\def\paperID{130} 

\title{RoomDesigner: Encoding Anchor-latents  for Style-consistent and Shape-compatible  Indoor Scene Generation}

\author{Yiqun Zhao$^1$ , Zibo Zhao $^1$, Jing Li $^{2}$ ,  Sixun Dong$^1$ , Shenghua Gao$^{1,3,4}$\footnote[1]{}\\
$^1$ShanghaiTech University \qquad $^2$Xiaohongshu Inc.\\
$^3$Shanghai Engineering Research Center of Intelligent Vision and Imaging\\
$^4$Shanghai Engineering Research Center of Energy Efficient and Custom AI IC\\
}

\begin{document}
\maketitle
\footnotetext[1]{Corresponding author}
\footnotetext[2]{contact: zhaoyq12022@shanghaitech.edu.cn}
\input{sec/0_abstract}    
\input{sec/1_intro}
\input{sec/2_relatedwork}
\input{sec/3_Method}
\input{sec/4_Experiments}
\input{sec/5_conclusion}

\clearpage
{
    \small
    \bibliographystyle{ieeenat_fullname}
    \bibliography{main}
}

\input{sec/6_suppl}

\end{document}

%% file: preamble.tex
\usepackage[dvipsnames]{xcolor}


%% file: sec/0_abstract.tex
\begin{abstract}


Indoor scene generation aims at creating shape-compatible, style-consistent furniture arrangements within a spatially reasonable layout. However, most existing approaches primarily focus on generating plausible furniture layouts without incorporating specific details related to individual furniture pieces. To address this limitation, we propose a two-stage model integrating shape priors into the indoor scene generation by encoding furniture as anchor latent representations.
In the first stage, we employ discrete vector quantization to encode furniture pieces as anchor-latents. Based on the anchor-latents representation, the shape and location information of the furniture was characterized by a concatenation of location, size, orientation, class, and our anchor latent. In the second stage, we leverage a transformer model to predict indoor scenes autoregressively. Thanks to incorporating the proposed anchor-latents representations, our generative model produces shape-compatible and style-consistent furniture arrangements and synthesis furniture in diverse shapes.
Furthermore, our method facilitates various human interaction applications, such as style-consistent scene completion,  object mismatch correction, and controllable object-level editing. Experimental results on the 3D-Front dataset demonstrate that our approach can generate more consistent and compatible indoor scenes compared to existing methods, even without shape retrieval. Additionally, extensive ablation studies confirm the effectiveness of our design choices in the indoor scene generation model.

 
\end{abstract}

%% file: sec/1_intro.tex
\section{Introduction}
\label{sec:intro}


\begin{figure}[htbp!]
    \centering
    \includegraphics[width=0.4\textwidth]{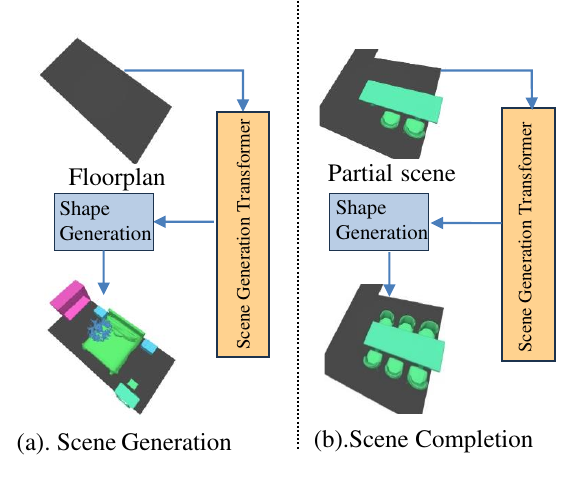}
     \vspace{-1em}
    \caption{Our proposed method can enable (a). Scene Generation (b). Scene Completion using the same Scene Generation Transformer followed by a furniture Shape Generation model. Unlike the previous retrieval-based method, our method can directly generate shapes for each piece of furniture.}
    \label{fig:teaser}
    \vspace{-1.5em}
\end{figure}

The automated generation of physically-plausible 3D indoor scenes is a challenging but important task in 3D vision. The capacity to digitally create immersive indoor environments holds immense potential across diverse industries, spanning gaming, architecture, virtual reality, simulation, and robotics. The scene generation process can be divided into two subproblems. The first one is to generate a plausible layout of the given scene. The second one is to obtain the style-consistent and shape-compatible furniture to decorate the indoor scene by following the generated layout. 

Lots of efforts have been made to generate physically-plausible indoor scenes~\cite{yang2021indoor, yang2021scene, paschalidou2021atiss, nie2022learning, Tang_2023}. Conventional scene synthesis methods \cite{merrell2011interactive, yeh2012synthesizing, chang2014learning, chang2017sceneseer, fisher2010context, fisher2015activity, fu2017adaptive} rely on predefined rules as constraints to guide the optimization process to achieve a reasonable synthesis result, but the design of rules highly depends on the knowledge from experts, which are not scalable. 
Recent advances in deep generative learning enable the synthesis of realistic indoor scene layouts~\cite{wang2018deep, li2019grains, wang2019planit, wang2021sceneformer, zhang2020deep, purkait2020sgvae, yang2021indoor, yang2021scene, paschalidou2021atiss, nie2022learning, Tang_2023} without reliance on predefined rules.
Specifically, most of these works first use a deep generative model to predict the layout of 3D/2D bounding boxes for the furniture elements within the scene, by leveraging GAN~\cite{yang2021scene}, VAE~\cite{purkait2020sgvae, yang2021indoor, wang2019planit, wang2018deep}, Autoregressive~\cite{wang2021sceneformer, nie2022learning, paschalidou2021atiss} or Diffusion model~\cite{Tang_2023}. Subsequently, the most suitable furniture shapes from a predefined CAD library are retrieved based on the predicted sizes of the bounding boxes.
Since the geometry and structure of furniture are complex, the commonly defined furniture parameters, including the size, color, translation, and rotation, are not sensitive to the furniture geometry and structure, consequently, directly retrieving the closest size furniture from the CAD library may lead to shape-incompatible and style-inconsistency issues with other furniture elements in the scene, for example, the chairs around the same table are not style consistent with each other, or the chair partially occupied the same space with the table (see the Fig. \ref{fig:Mainexperimets} in the experiments).
 Further, the retrieval-based solution cannot generate novel shapes because of the constraint of the pre-defined CAD library.  
To avoid this issue, a very recent work~\cite{Gao2023scenehgn} proposes to use the part-level embedding combined with a graph representation to model the geometry and structure of furniture in the scene. However, their work highly relies on the part-level annotation of objects, which is usually unavailable, consequently restricting the application of these approaches on large-scale dataset like 3D-Front dataset~\cite{fu20213d, fu20213dfuture}.
 
 In light of the recent successes in high-quality 3D shape generation~\cite{hui2022neural, zeng2022lion, zhao2023michelangelo, zhang20233dshape2vecset, Zhou_2021_ICCV}, we propose to generate both scene layout and 3D furniture shapes, simultaneously. Towards this end, a suitable shape representation that captures both the global structure and the local geometry details of the furniture is highly demanded. Besides, the shape representation should be easily scaled up to the scene level, which is composed of many different pieces of furniture. 
 Inspired by the recent works \cite{zhang2022dilg, esser2021taming}, we propose a hybrid representation for scene generation, where we use anchor points sampled from the dense point clouds for the global structure representation, which is computatinally efficient. We further propose to extract the local latent features of the anchor points by the VQ-VAE~\cite{zhang2022dilg}, which encodes the surface points sampled from discrete mesh into the latent representation. Therefore, the latent representation of the anchor points contains detailed information. Then we leverage a decoder to decode our latent anchor representation into an occupancy field. With the local latent features, the anchor-points can generate a smooth surface and provide the geometry details. 

 We parameterize the furniture with the anchor-latents, spatial location, size, rotation, and category. Since auto-regressive modeling with a transformer is the most straightforward way to make the furniture generation conditional on all previous furniture existing in the scene. We then learn a transformer-based architecture to model the process of scene generation as the next-token prediction within an unordered set. The parameters of each piece of furniture in the scene are designed to be both geometry- and structure-aware, allowing for a comprehensive understanding of the objects’ spatial characteristics. As a result, our approach has demonstrated significant improvements in style consistency and shape compatibility among the generated furniture items. Our proposed geometry and structure-aware anchor-latents not only improve the room-mask-conditioned scene generation but also benefit a lot of human interaction applications that should take geometry into consideration, such as style-consistent scene completion, furniture mismatch correction, and controllable object-level scene editing. Furthermore, the introduced anchor-latents facilitate geometry generation modeling, empowering our model to generate novel shapes for each furniture category. Extensive experiments show that our method achieves compelling results in the room-mask conditional generation settings against state-of-the-art scene generation models. 

\medskip
\noindent
 Our contributions can be summarized as follows:
\begin{itemize}
    \item We introduce the anchor-latents to encode the geometry and structure of each piece of furniture to achieve consistent and compatible indoor scene generation without any part-level annotations.
    \item Based on the proposed anchor-latents, we learn a geometry-aware scene generation transformer. It cannot only enable style-consistent and shape-compatible but also facilitates downstream tasks such as style-consistent scene completion, object mismatch correction, and controllable object-level scene editing.
\end{itemize}





%% file: sec/2_relatedwork.tex
\section{Related Work}

\noindent\textbf{Indoor scene generation.} 
Most existing Indoor scene synthesis methods can be divided into two main different parts. On the one hand, traditional scene synthesis methods~\cite{merrell2011interactive, yeh2012synthesizing, chang2014learning, chang2017sceneseer, fisher2010context, fisher2015activity, fu2017adaptive} mainly rely on human-defined constraints as scene priors to guide the optimization process. While these works can produce some reasonable results, the results of such methods are limited to simple patterns. On the other hand, recent advances in deep generative learning ~\cite{wang2018deep, li2019grains, wang2019planit, wang2021sceneformer, zhang2020deep, purkait2020sgvae, yang2021indoor, yang2021scene, paschalidou2021atiss, nie2022learning, Tang_2023, CC3D, Xu_2023_CVPR} has greatly facilitated the indoor scene synthesis. These work explore different modeling such as VAEs~\cite{wang2018deep, li2019grains, wang2019planit, purkait2020sgvae, yang2021indoor}, GANs~\cite{yang2021scene, CC3D, Xu_2023_CVPR}, Autoregressive~\cite{paschalidou2021atiss, nie2022learning, wang2021sceneformer} and Diffusion model for scene synthesis. Closely related to our work are autoregressive indoor scene generation models~\cite{wang2018deep, wang2019planit, paschalidou2021atiss, nie2022learning, wang2021sceneformer}. ATISS\cite{paschalidou2021atiss}~propose an unordered set representation for auto-regressive scene synthesis. Each piece of furniture is represented as the category, location, orientation, and size in their work. This can achieve a more robust scene-generation process. However, it is still not sensitive to furniture shape. Scene Priors~\cite{nie2022learning} inject the shape embedding for each piece of furniture in their model. They defined a template mesh~\cite{wang2018pixel2mesh} for each piece of furniture and predicted the deformation for the template vertex of each piece of furniture. The optimization process was achieved through differentiable rendering~\cite{lassner2020pulsar} using multiview images. Due to the limited topology of the template mesh, it cannot directly decode reasonable shapes. Some works\cite{CC3D,Xu_2023_CVPR} explore volume rendering\cite{mildenhall2020nerf} for scene synthesis supervised by multiview images. However, this line of work can not produce high-quality geometry for each piece of furniture. Most recently, SceneHGN~\cite{Gao2023scenehgn} proposed to use a hierarchical VAE to encode the scene at part-level. It can produce a reasonable shape for each piece of furniture, but their method highly relies on part-level annotation.



\noindent\textbf{3D shape representation.} 
3D shape representation has been explored for many years~\cite{chen2018implicit_decoder, 3dgan, zhang20233dshape2vecset, autosdf2022, mildenhall2020nerf, cheng2023sdfusion, Mescheder2019CVPR, Peng2020ECCV, gao2022get3d, gupta20233dgen, shen2021dmtet, Liu2023MeshDiffusion, zhang2022dilg, liu2023openshape, xue2022ulip, xue2023ulip2}. Representing the 3D Shape~\cite{ Tang_2023} into a 1d-vector is compact but may lead to the lack of geometry details. Recently, OccNet~\cite{occnet} proposed to use a dense feature voxel to represent the 3d shape. While the dense voxel is too dense,  some other works~\cite{Peng2020ECCV, gupta20233dgen, Chan2021, gao2022get3d}  propose to decompose the 3d voxel into a tri-plane representation, where the complexity efficiency can be reduced. Some other interesting works propose to decompose the 3D feature volume into a low-rank~\cite{Chen2022ECCV, gao2023strivec} representation. All the existing local representation is computationally expensive, the expensive computation makes them hard to scale up to the scene level. Some works~\cite{autosdf2022, Mescheder2019CVPR, liu2023openshape, xue2022ulip} propose to encode the shape as a single latent vector, which can provide high-level structure information. But the global representation is insufficient to capture the geometry details. It is desirable to represent the scene with both global structure and local details. Due to the sparse nature of 3D shapes themselves, another line of work proposes to use the irregular latent fields~\cite{yan2022shapeformer,zhang2022dilg} to represent each shape in a more efficient way.  Inspired by this previous line of work, we propose anchor-latents representation for scene generation task. 


%% file: sec/3_Method.tex
\section{Method}



To generate shape-compatibility and style-consistency furniture within a room mask obtained from a top-view perspective projection from an empty or partially decorated room, we devise an anchor-latent representation and a geometry-aware scene generator. The generator captures the geometry information of the furniture benefiting from the anchor-based latent representation. It then autoregressively predicts shape-compatible and style-consistent furniture by injecting location information from the furniture. 

Specifically, our proposed two-stage generative model consists of two modules, a Vector Quantised-Variational AutoEncoder (VQ-VAE) for representing the furniture into anchor-latents representation (see Section~\ref{subsec:anchor}) and a geometry-aware transformer that produces indoor scenes autoregressively, which composes with a scene generation transformer followed by a shape generation branch based on the proposed anchor-latents (see Section~\ref{subsec:generation}).

\begin{figure*}[!htbp]
    \centering
    \includegraphics[width=0.98\textwidth]{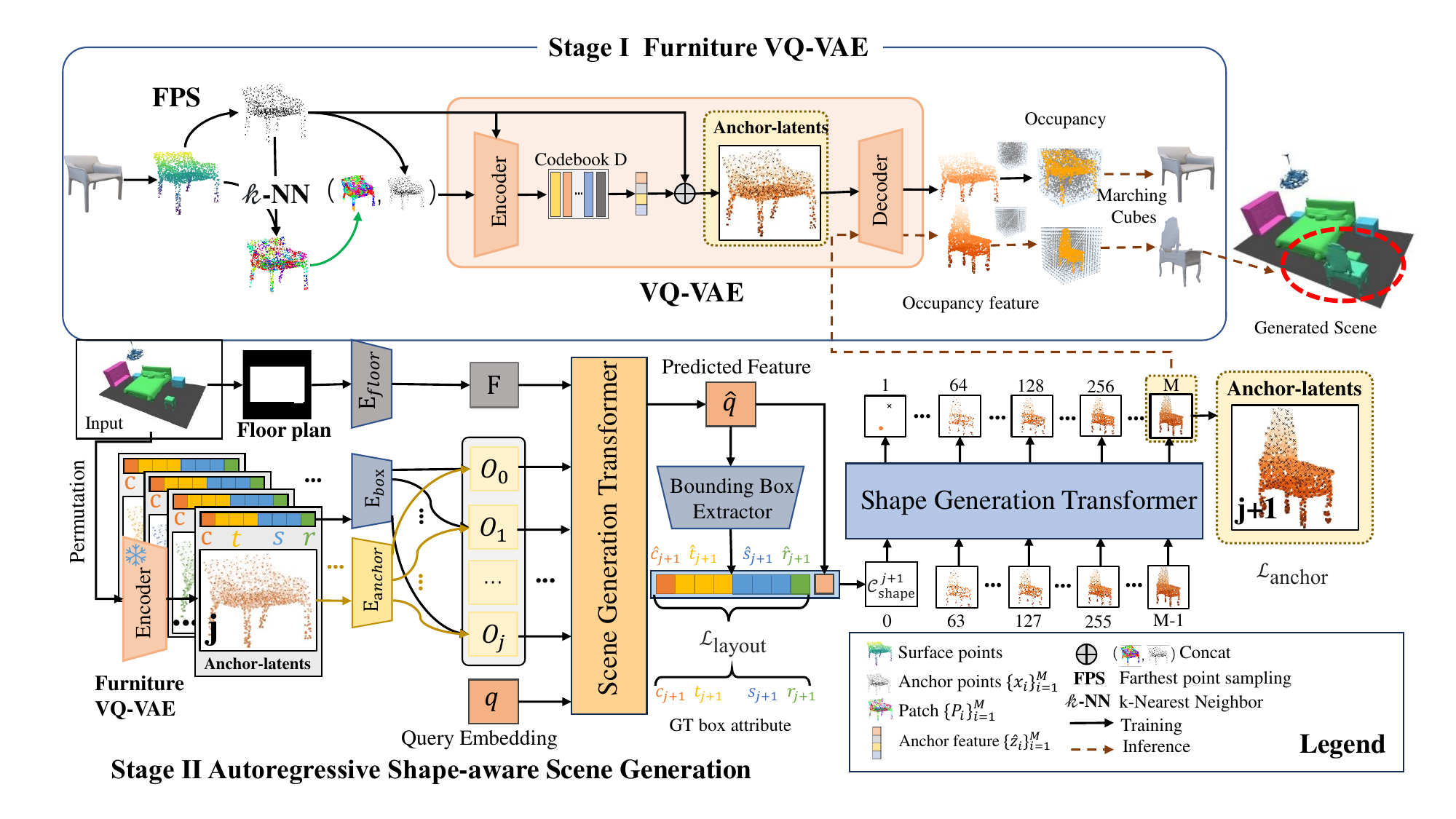}
    \vspace{-0.5em}
    \captionsetup{belowskip=-6mm}
    \caption{Our proposed two-stage method. In the first stage, We sample the surface points from the furniture surface and extract anchor points from surface points using FPS, then use $k$-NN to obtain the patch. We then feed them into a VQ-VAE to get the anchor-latents representations. The VQ-VAE is supervised by decoding the anchor-latents into an occupancy field. In the second stage, given a scene with $j$ objects and a room mask, the floor plan encoder maps the floor into a feature $F$,  the box encoder and anchor encoder map the box attribute and furniture anchor-latents into a per-furniture shape-aware embedding  $\{O_i\}_{i=1}^j$. The floor plan features $F$, the shape-aware embedding $O$, and a learnable query embedding are then fed to the scene generation transformer that predicts $\hat{q}$. Using $\hat{q}$, we first extract the bounding box of all furniture for layout synthesis. We then concatenate the box attribute with $\hat{q}$ as a condition for shape autoregressive generation. We jointly train the scene generation transformer for layout synthesis and the shape generation transformer for furniture generation. During the inference stage,  The input can be only the floor plan (for scene generation) or the floor plan with partial furniture (for scene completion).}  
    \label{fig:method}
\end{figure*}

\subsection{Anchor-latents representation}

\label{subsec:anchor}

It is challenging to achieve learning-friendly 3D shape representation due to its high-dimensional nature. Especially in scene composition modeling, since a scene may be composed of more than ten objects. Traditional shape representation methods~\cite {Peng2020ECCV, Mescheder2019CVPR} are still too dense to represent the shape. To make the learning of 3D shape distributions in an efficient way while also keeping its geometry information in this representation, we compress each piece of furniture into the anchor-latent representation. It is composed of a set of anchor points $\{\mathrm{x_i}\}_{i=1}^M, \mathrm{x_i}\in \mathbb{R}^3$ and its corresponding latents $\{ \hat{z}_i \}_{i=1}^M, z_i\in \mathbb{R}^{C}$, where $M=512$ is the number of anchor points and $C$ is the feature channel dimension. Compared with the previous shape representation method, such a representation is more lightweight. It can also preserve high-quality geometry information in it. We achieve this anchor-latents encoding by training a Furniture VQ-VAE~\cite{esser2021taming, zhang2022dilg} for occupancy fileds, as shown in Fig.~\ref{fig:method}

\noindent\textbf{Anchor-point extraction.} Specifically, given a watertight mesh , we random sample $N$=2048 points from its surface. The goal of this stage is to predict the occupancy indicator $O(\cdot):\mathbb{R}^3\rightarrow [0, 1]$ from the randomly sampled $N$ points. To build the anchor-latens representation, We first use FPS (Farthest Point Sampling) to get the $M$ anchor points. 

\noindent\textbf{Anchor-feature extraction.} With the point set $\{\mathrm{x_i}\}_{i\in M}$ as center points, we use $k$-NN ($k=32$) to get $k$-1 nearest neighbor points from the initial N surface points for each anchor point to form a local patch $\{P_i\}_{i=1}^M $. We then feed each local patch and anchor-points into a VQ-VAE Encoder to extract the anchor-feature $z_i$
\begin{equation}
\{z_i\in \mathbb{R}^C\}_{i=1}^M = \text{Encoder}(\{{(x_i, \mathrm{P_i})}\}_{i=1}^M)
\end{equation}


We then use the vector quantization~\cite{oord2017neural} to translate the $z_i$ into $\hat{z_i}$ from a codebook $D$ ($|\mathcal{D}| = 1024$),

\begin{equation}
    i = \text{argmin}_{\hat{z}_i \in D} ||\hat{z_i} - z_i ||
\end{equation}

To optimize the anchor-latents $\{(\mathrm{x_i}, \hat{z}_i)\}_{i\in M}$, we feed them into a transformer-based decoder to get the occupancy fields feature. To get the final occupancy prediction, we feed the anchor-latents into a decoder to get the Occupancy feature for each query point $\mathrm{v}$. We calculate the interpolated feature based on the distance between query point $\mathrm{v}$ and our anchor points $\{\mathrm{x_i}\}_{i=1}^M$. Finally, the interpolated feature $z_v$ was fed into an MLP to get the occupancy prediction $\hat{O}(\mathrm{v})$. 

The anchor-latents learning stage is supervised by comparing the predicted occupancy to the ground truth occupancy for the query points $v$ with a binary cross entropy loss. An additional commitment loss is used to regularize the codebook. The overall loss for this stage is:
\begin{equation}
    \mathcal{L} = \mathcal{L}_{\text{occ}} + \lambda \mathcal{L}_{\text{commit}}
\end{equation}\begin{equation}
    \mathcal{L}_{\text{occ}} = \mathbb{E}_{\mathrm{v}\in \mathbb{R}^3}\left[\text{BCE}(\hat{O}(v), O(v))\right] 
\end{equation}\begin{equation}
    \mathcal{L}_{\text{commit}} = \mathbb{E}_{\mathrm{v}\in \mathbb{R}^3 }\left[\mathbb{E}_{i \in M } \| \text{sg}(\hat{z}_i )- z_i\|^2\right]
\end{equation}

\subsection{Autoregressive Shape-aware Scene Generation}

\label{subsec:generation}

After obtaining the anchor-latents of each object, we feed the shape anchor-latents and the bounding box locations into a Scene Generation transformer. The generation process is achieved by Auto-regressive modeling, as shown in the second stage of Fig.~\ref{fig:method}.

\noindent\textbf{Auto-regressive scene generation modeling.} The scene $\mathcal{S}$ is composed of floor plan $\mathcal{F}$ and objects $O = \{o_i\}_{i=1}^L$, where L is the number of objects in the scene.  Given a floor plan $\mathcal{F}$ for a scene $\mathcal{S}$, our goal is to generate shape-compatible and style-consistent objects $O$. We parameterize the object as a concatenation of category $c_i$, size $s_i \in \mathbb{R}$, rotation $r_i\in \mathbb{R}^3$, translation $t_i \in \mathbb{R}^3$ and our introduced anchor-latents $\{(\mathrm{x_i}, \hat{z}_i)\}_{i\in M}$. Our goal is to model the distribution over scene $p_{\theta}(S)$, which can be written as 

\begin{equation}
\footnotesize{
    P_{\theta}(\mathcal{S}) = P(\mathcal{F}) P_{\theta}(\mathcal{O} | \mathcal{F}) = P(\mathcal{F}) \sum_{o \in \pi(O)} \Pi_{j \in o} P(o_j| o_{<j}, \mathcal{F})
}
\end{equation}
where $P(o_j| o_{<j}, \mathcal{F})$ is the probability of the $j$-th object, conditioned on the previously generated objects and the floor layout, and $\pi(\cdot)$ is a permutation operator. The objective is to maximize the log-likelihood of this function. 

\begin{equation}
    \log P_{\theta}(\mathcal{S}) =   \sum_{o \in \pi(O)} \sum_{j \in o} \log P_{\theta}(o_j | o_{<j}, \mathcal{F})
\end{equation}

In order to make our model invariant to the order of generated objects, we follow~\cite{paschalidou2021atiss} to apply permutations on the object sequence during training. This can facilitate a lot of downstream tasks, such as scene completion. 

\noindent\textbf{Floor-plan Encoder.} The 2D room mask for the scene shape is encoded with a ResNet-18~\cite{he2016resnet}. The encoded feature $F$ provides information to guide the model in predicting a reasonable location of the furniture in an empty space of the room.

\noindent\textbf{Bounding Box Encoder.} We encoded the category $c$ of each bounding box using a learnable embedding $\theta_c(\cdot)$. A positional encoding $\text{pos}(\cdot)$~\cite{vaswani2017attention} is used for the translation $t$, rotation $r$ and size $s$.
\begin{equation}
\footnotesize{
E_{\text{box}}(c_j, t_j, r_j, s_j) \rightarrow (\theta_c(c_j), \text{pos}(t_j), \text{pos}(r_j), \text{pos}(s_j))
}
\end{equation}

\noindent\textbf{Anchor-latents Encoder.} For each object, we have anchor-latents as $\{(x^i, \hat{z}^i)\}_{i=1}^M$,  as described in Sec.\ref{subsec:anchor}. However, the dimension of the anchor-latents, $M \times (C+3) $= $512 \times (256+3)$ (here $C=256$), is still too high to be directly injected with the bounding boxes tokens. For more efficiency, we use the quantized index $\{id^i_j\}_{i=1}^M$ of each anchor feature in the codebook to represent the feature from our codebook. Therefore, the total dimension for shape embedding can be reduced to $M\times 4$. For each anchor in the anchor groups, we directly map them with a positional encoding $\text{pos}(\cdot)$ to map them as $a \in \mathbb{R}^{M\times T}$, where T is the embedding dimension of positional embedding,  then a learnable embedding $\theta_{\text{anchor}}$ is used to embed it into a 1D vector $g \in \mathbb{R}^M$.
\begin{equation}
\vspace{-0.3em}
    E_{\text{anchor}} : (x_j^i, id_j^i) \rightarrow \theta_{\text{anchor}} (\text{pos}(x_j^i, id_j^i))
\end{equation}
We then concatenate the $g$ with the bounding box embedding to get the per furniture parameters $O_j$.

\noindent\textbf{Scene Generation Transformer.} We feed the room-mask feature $F$, per-furniture context embeddings $O_{<j}$, and a query embedding $q$ into the scene transformer encoder. It predicts the furniture features $\hat{q}$ for the subsequent object generation. 
\begin{equation}
\vspace{-0.3em}
    \tau_{\text{scene}}(F, O_{<j},q) \rightarrow \hat{q}
\end{equation}

\noindent\textbf{Bounding boxes attribute extractor.} For bounding boxes attribute distribution $(\hat{c}, \hat{t}, \hat{r}, \hat{s} )$ decoding from $\hat{q}$, we follow the same design from ATISS~\cite{paschalidou2021atiss}. To be more specific, we employ an MLP for each attribute in a sequential manner. That is, given $\hat{q}$, we first predict the category label $\hat{c}$. Then we predict the following attribute $\hat{t}$, $\hat{r}$, $\hat{s}$, sequentially. The predicted previous attribute will be concatenated with the $\hat{q}$ for the next attribute prediction. This means that the next attributed prediction is based on the previous attributed prediction results. Based on this property, we can extend the shape model as shape generation in a similar way. 

\noindent\textbf{Shape generation Transformer.} After bounding boxes attribute prediction, we can get a concatenate of predicted attribute $\mathcal{C}_{\text{shape}}=(\hat{q}, \hat{c}, \hat{t}, \hat{r}, \hat{s})$,  we generate the anchor-latens for the next object conditional on all these predicted attributes. We use a transformer to achieve anchor-latents autoregressively generation. Thanks to the modeling of the distribution of anchor points, our shape branch can be easily extended to object-level editing by controlling these anchor points. Specifically, we predict $M$ anchor-latents, sequentially. These latents are re-ordered in ascending order by anchor-points coordinates. This can ensure our generation process follows a specific order. 
Speacifically, the anchor-latents $(x_i, z_i)$ is conditional on all previous prediction anchor-latents $(x_{<i}, z_{<i})$ This anchor distribution can be written as:
\begin{equation}
\scriptsize{
\begin{aligned}
P( \{(x_i, z_i)\}_{i=1}^M| \mathcal{C}_{\text{shape}} ) =
\Pi_{i=1}^M P((x_i, z_i)| \{(x_j, z_j)\}_{j=0}^{i-1}, \mathcal{C}_{\text{shape}} )
\end{aligned}
}
\end{equation}

\subsection{Training}

To train our model, We randomly select $j$-1 object in a scene and apply a random permutation to it. We first use the room mask Encoder to get the room mask feature $F$. We then use the bounding box encoder and anchor-latents encoder to get the corresponding embedding of the $m$ pieces of furniture $O_{<j}$. We then feed them into the scene generation transformer to get the prediction token for the next object $\hat{q}$. This $\hat{q}$ is conditional on the information from both room layout and previous m bounding boxes and furniture geometry. We feed the $\hat{q}$ in bounding boxes attribute extractor and shape generation branch sequentially to get both bounding box attribute  $(\hat{c}, \hat{t}, \hat{r}, \hat{s})$ and each furniture anchor-latents $\{(x^i, \hat{z}^i)\}_{i=1}^M$. Since our model is fully auto-regressive based solution, we can train our model using by maximizing the log-likelihood of bounding box attributes $(\hat{c}, \hat{t}, \hat{r}, \hat{s})$ using $\mathcal{L}_{\text{layout}}$ and furniture anchor-latents $\{(x^i, \hat{z}^i)\}_{i=1}^M$ using $\mathcal{L}_{\text{anchor}}$. Different from previous work\cite{nie2022learning} that use a two stage training for layout and shape. We jointly train the layout and shape generation. We conduct ablation study to verify this. The implementation details can be found in the supplementary.

Once the model is trained, we sequentially generate each piece of furniture based on the room mask or the partially observed scene. 

%% file: sec/4_Experiments.tex
\section{Experiments}

\noindent\textbf{Datasets.} For experimental comparisons, we conduct our experiments on the large-scale 3D indoor scene dataset 3D-Front~\cite{fu20213d} furniture with 3D-Future~\cite{fu20213dfuture} CAD model. 3D-Front is a synthetic dataset composed of 6,813 houses with 14,629 rooms, where each room is arranged by the furniture from the 3D-Future dataset~\cite{fu20213dfuture}. Following the filtering strategy of ATISS~\cite{paschalidou2021atiss}, we use 4,041 bedrooms, 900 dining rooms, and 813 living rooms. There are 16,565 3D objects in 3D-Future~\cite{fu20213dfuture}  datasets. We selected 9,495 3D data from the 3D-Future dataset~\cite{fu20213dfuture} for shape autoencoder training. These 3D objects belong to the furniture categories present in our filtered scenes. Note that for all room types, we use the same shape auto-encoder.

\noindent\textbf{Baselines.} To verify the efficiency of our method, we qualitatively and quantitatively evaluate our method and compare it with the two most recent open-sourced baselines with the same experiment settings, including an autoregressive-based method ATISS ~\cite{paschalidou2021atiss} and VAE-based method Sync2Gen~\cite{yang2021scene}\footnote{Since Sync2Gen~\cite{yang2021scene} are designed for unconditional synthesis, for fair comparison, we replace the VAE as conditional VAE~\cite{NIPS2015_8d55a249} conditioned on the room mask.}. We also design another baseline by injecting the state-of-the-art pre-trained shape embedding OpenShape~\cite{liu2023openshape} into ATISS and retrieve the object from the CAD library using both the object size and OpenShape~\cite{liu2023openshape} embeddings in the inference stage.


\noindent\textbf{Evaluation Metrics.} To measure the quality and plausibility of our generated results, following previous works~\cite{nie2022learning, paschalidou2021atiss, wang2021sceneformer}, we use Frechet inception distance(FID) ~\cite{heusel2017gans}, scene classification accuracy(SCA) and category KL divergence of our 1,000 generated scenes. For FID, SCA, we render the generated scenes and ground-truth scenes from the test set into 256$\times$ 256 images through top-down orthographic projections. Following previous work~\cite{nie2022learning}, the color of each object is determined by the specific color related to the category. We also report the KL divergence between the object category distributions of synthesized and real scenes from the test set. To measure the layout plausible and shape-compatible, we use a collision rate to measure the proportion of collision objects generated among all the generated objects within a scene. 
To measure the shape consistency, we mainly consider dining chairs for the dining room and living room. The dining chair within a single scene should be consistent with each other. Specifically, we use OpenShape~\cite{liu2023openshape} similarity score to evaluate the semantic consistency and Chamfer-Distance to evaluate the detailed consistency. Finally, to measure the shape diversity of our generated shape, we directly run ten times conditioned on the same room mask and calculate the Chamfer-Distance of the same category object in different runs. We mainly consider the two classes of objects in each room type.


\subsection{Room mask conditioned scene generation.}

\begin{figure*}[!htbp]
    \centering
    \includegraphics[width=0.99\textwidth]{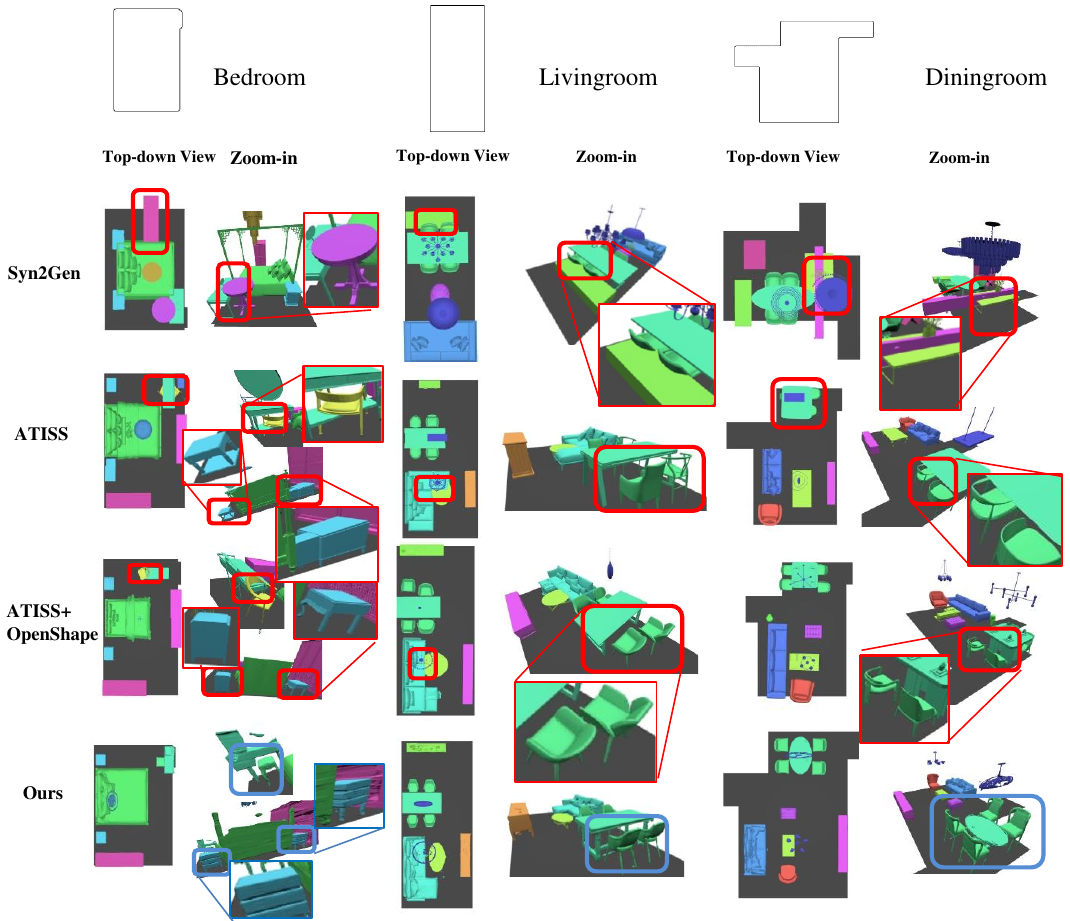}
    \vspace{-4mm}
    \caption{Scene Generation conditioned on room mask. Note that all the baselines synthesize the indoor scene with the retrieved furniture from a predefined CAD library. Our method directly generates all the furniture from our generation model. The red rectangles highlight the style inconsistency and shape incompatible with the baseline results. The blue rectangle highlights that the rooms generated by our method can ensure shape-compatibility and style consistency.}
    \label{fig:Mainexperimets}
    \vspace{-1em}
\end{figure*}

Fig.~\ref{fig:Mainexperimets} visualizes the qualitative comparisons of different scene synthesis methods. We can see that baseline methods cannot guarantee style consistency and shape compatibility, especially for the dining room case where exist usually multiple dining chairs. However, our method improves both. Further, there are mainly two reasons that lead to collisions occurring, one is the layout prediction error, and the other is the shape-incompatible between different scenes (e.g., chairs under the same table should be compatible with each other). It shows that  with our proposed anchor-latents, the generation process of our method will be shape-aware. Tab.~\ref{tab:roommask} further proves that our method can produce high-quality while also keeping the style consistent and shape-compatible results, especially for the dining room. 

\begin{figure}
    \centering
    \includegraphics[width=0.45\textwidth]{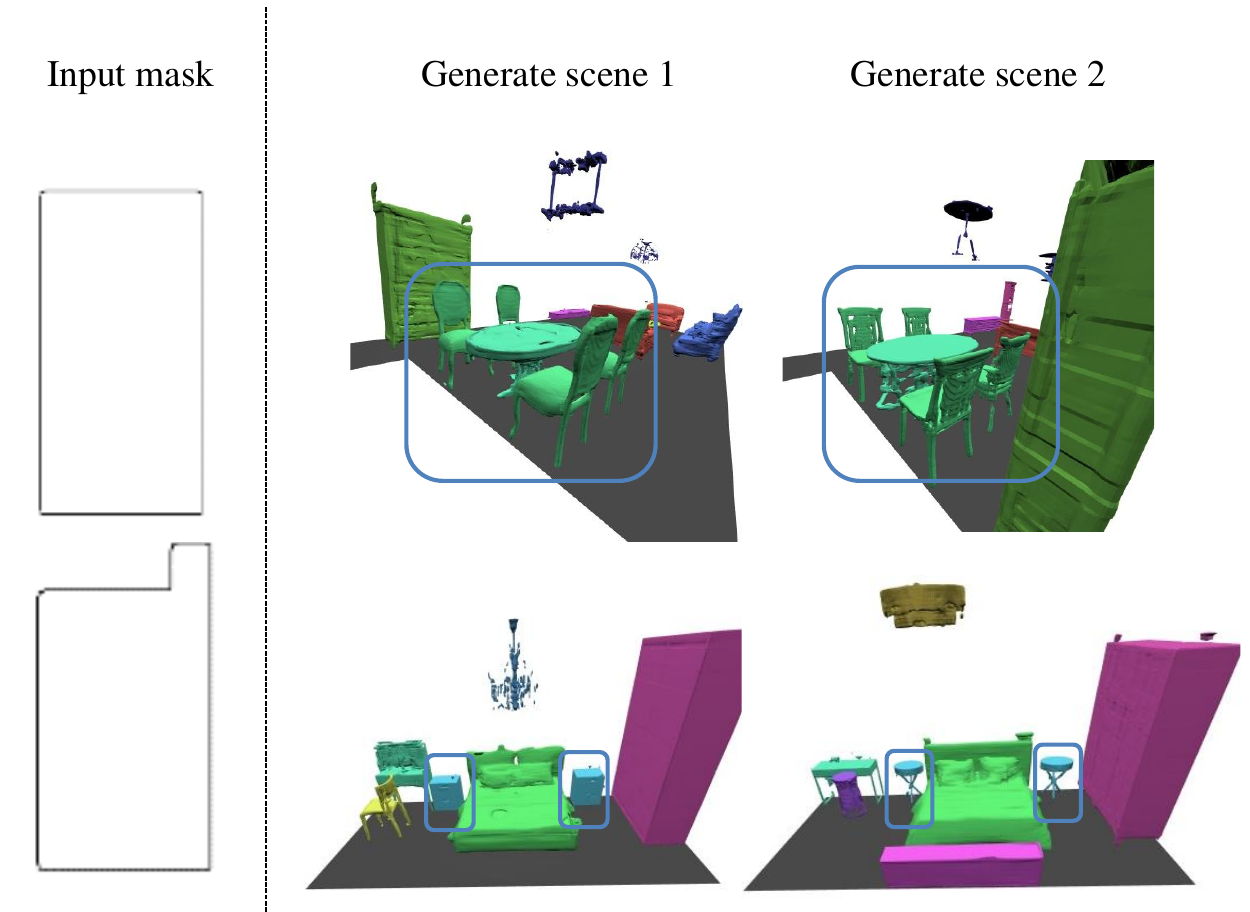}
     \vspace{-3mm}     
    \caption{Consistency and Diversity of our results. We run the experiments conditional on the same floor plan and visualize the results from different runs. Our method can preserve the same category consistent within each generated scene. For different  generated scenes, furniture from the same category are different .}
    \vspace{-1.5em}  
    \label{fig:diversity}
\end{figure}

\noindent\textbf{shape-consistency and shape-diversity.}
Fig.~\ref{fig:diversity} visualizes the qualitative results of our method when given the same room mask. We can observe that for the same object category, our method can generate style-consistent furniture within a scene while can generate diverse furniture. Tab~\ref{tab:consistency} and  Tab~\ref{tab:diversity} further prove that. In Tab.~\ref{tab:consistency},  ATISS\cite{paschalidou2021atiss} + OpenShape~\cite{liu2023openshape} shares almost similar O-score with ours, which means that both the OpenShape~\cite{liu2023openshape} embeddings and our proposed anchor-latents can provide sufficient global structure information. But when compared the Chamfer-Distance, our method outperforms it significantly. This indicates that our proposed anchor-latents can capture both global information and geometry details for the furniture.

\setlength{\tabcolsep}{4pt}
\begin{table}[!hbt]
        \centering
	\renewcommand\arraystretch{1.2}
	\begin{center}
 \resizebox{.48\textwidth}{!}{
		\begin{tabular}{*{13}{c}}
			\toprule
			\multirow{2}*{Method}  & \multicolumn{4}{c}{Bedroom}   & \multicolumn{4}{c}{Dining room}  & \multicolumn{4}{c}{Living room} \\
			\cmidrule(lr){2-5} \cmidrule(lr){6-9} \cmidrule(lr){10-13}
			 & FID $\downarrow$  & SCA $\%$ & CKL $\downarrow$& Colli $\downarrow$    & FID $\downarrow$  & SCA $\%$ & CKL $\downarrow$ & Colli $\downarrow$  & FID $\downarrow$ & SCA  $\%$ & CKL $\downarrow$   & Colli $\downarrow$ \\ 
			\midrule
			\midrule


            Sync2Gen*~~\cite{yang2021scene}
                  & 49.32  & 79.63  & 0.031  & 0.466
                  & 57.44  &  82.11 & 0.092  & 0.442 
                  & 59.94 &  88.40 & 0.063 & 0.498
                  \\

            Sync2Gen~~\cite{yang2021scene}
                  & 46.16 & 83.12  & 0.043 & 0.344
                  & 59.97 &  84.32 & 0.084  & 0.427
                  & 60.96 &  86.25 & 0.074 & 0.477 \\

            ATISS~~\cite{paschalidou2021atiss}
                  & 29.78  & 67.37 &  0.011 &  0.326 
                  & 37.37 &  79.44 &  0.006 &   0.447
                  & 32.10  & 81.25 & 0.005 &  0.439 \\

            ATISS~\cite{paschalidou2021atiss}+OpenShape~\cite{liu2023openshape}
            & 28.86  &  64.74 & \textbf{0.009}   & 0.317
            & 33.69  & 69.81 & \textbf{0.003}  &   0.452 
            & 31.59  &  72.57 & 0.006   & 0.436 \\

            Ours
                 & \textbf{27.16} & \textbf{63.58} & \underline{0.010}  & \textbf{0.287} 
                   & \textbf{32.78} & \textbf{65.53} & \textbf{0.003}  & \textbf{0.360} 
                    & \textbf{30.88} & \textbf{69.92} & \textbf{0.004}  & \textbf{0.371}  \\

        \bottomrule
        \end{tabular}
         }
         \vspace{-0.5em}
        \caption{Quantitative comparisons on the task of room-mask conditional generation. Sync2Gen* is the version of ~\cite{yang2021scene} without Bayesian optimization. FID, SCA is calculated by render the scene from top view into a 256$\times$ 256 resolution image. CKL means the category distribution KL divergence between the test set and our generated set. Colli means the average collision rate of the generated object in the generated scenes. For FID  CKL and Colli, the lower means the better. SCA closer to 50\% is better.}
        \label{tab:roommask}
       
        \end{center}
        \vspace{-8mm}
\end{table}

%

\setlength{\tabcolsep}{2pt}
\begin{table}[!hbt]
    \renewcommand\arraystretch{1.2}
    \begin{center}
    \resizebox{0.35\textwidth}{!}{
        \begin{tabular}{*{7}{c}}
            \hline
            \multirow{2}*{Method}  & \multicolumn{2}{c}{Dining room}  & \multicolumn{2}{c}{Living room}  \\
            \cmidrule(lr){2-3} \cmidrule(lr){4-5} 
             & CD $\downarrow$ & O-score $\uparrow$ & CD  $\downarrow$  & O-score $\uparrow$ &
     \\ 
            \hline\hline
           
            ATISS~~\cite{paschalidou2021atiss}
                  & 93.0  & 0.92 &  30.0 &    0.91 \\

            ATISS~\cite{paschalidou2021atiss}+ OpenShape~\cite{liu2023openshape} 
            & 12.1 &    \textbf{0.98} & 12.6 & \textbf{0.97}
            \\
             
            Ours
                 & \textbf{6.9} & \textbf{0.98}  & \textbf{9.7} & \textbf{0.97}\\
        \hline
        \end{tabular}
        }
        \vspace{-2mm}
        \caption{Quantitative comparisons on shape consistency for room-mask conditional generation. CD means Chamfer-Distance($\times$ 0.001),  the lower means the geometry details between two pieces of furniture are more similar. O-score means OpenShape~\cite{liu2023openshape}  similarity score. The higher O-score, the more semantic similar the two pieces of furniture are.}
        \label{tab:consistency}
        \end{center}
        \vspace{-2em}
\end{table}


\setlength{\tabcolsep}{3.5pt}
\begin{table}[!htbp]
	\renewcommand\arraystretch{1.2}
	\begin{center}
 \resizebox{0.35\textwidth}{!}{
	\begin{tabular}{*{7}{c}}
            \toprule
            \multirow{2}*{Method} & \multicolumn{2}{c}{Bedroom}   & \multicolumn{2}{c}{Dining room}  & \multicolumn{2}{c}{Living room}  \\
            \cmidrule(lr){2-3} \cmidrule(lr){4-5} \cmidrule(lr){6-7}
             & Bed $\uparrow$ & Nightstand $\uparrow$ & Chair $\uparrow$  & Table $\uparrow$ & Chair $\uparrow$  & Table $\uparrow$
     \\ 
            \midrule
            \midrule
            
            ATISS~~\cite{paschalidou2021atiss}
                 & 1.89  & 0.93 &  
                 2.45 &  1.92 & 3.29 & 2.66  \\
            ATISS~\cite{paschalidou2021atiss}+ OpenShape~\cite{liu2023openshape} 
            &  2.56 & 1.42 & 7.62
            & 4.34  & 6.40 & 3.34
            \\
            Ours
                 & \textbf{25.4} & \textbf{7.3} & \textbf{29.4} 
                   & \textbf{17.7} & \textbf{22.3}  & \textbf{16.8}  \\
        \bottomrule
        \end{tabular}
        }
        \vspace{-2mm}
        \caption{Quantitative evaluation on the shape diversity when given the same scene mask with 10 run times. We choose 2 classes in each room type and report the Chamfer-Distance($\times 0.001 $) here.} 
        \label{tab:diversity}
        \end{center}
        \vspace{-5mm}
\end{table}

\subsection{Ablation studies}
We ablate our method with different numbers of our latent points and training strategies for our scene-generation transformer under room-mask-conditioned generation settings. The results are shown in Tab. \ref{tab:ablation}.

\noindent\textbf{Number of anchor points.}  We use different anchor points to conduct experiments on the Bedroom of the 3D-Front dataset. Since more anchor points can provide more global structure information for each piece of furniture, with very few global structure information, Our scene generation transformer will not be shape-aware, where collision rate can prove this. When using only 128 anchor-latents, the collision rate is even worse than  methods that do not consider geometry. Furthermore, fewer anchor points can not provide reasonable geometry for each piece of furniture, thus leading to a decrease in FID score and SCA. Please see supplementary for qualitative results.

\noindent\textbf{The training strategy.} We ablate our Scene Generation Transformer training strategy with a two-stage training strategy proposed by Scene Prior~\cite{nie2022learning}. At the first training stage, the shape generation transformer of our model was not engaged in training. Only class category and bounding boxes parameterization was optimized. Then we jointly optimize the geometry branch and train our model end to end. We have observed a moderate decline in both the FID and SCA. However, the reduction of CKL is serious. The gains can be attributed to the joint training of scene and shape, which facilitates the learning of category distribution since geometry distribution varies for each category.

\setlength{\tabcolsep}{3.5pt}
\begin{table}[!htbp]
	\renewcommand\arraystretch{1.2}
	\begin{center}
		\begin{tabular}{*{5}{c}}
			\hline
                Method & FID $\downarrow$& SCA $\%$  & CKL $\downarrow$  & Colli $\downarrow$ \\ 
            \hline\hline
            Ours-128
                   & 33.68 &  81.24 & 0.019   & 0.323 \\

            Ours-256
                  & 29.96 &  70.70 &  0.011   & 0.314 \\
            \hline
            Two-stage
                &  28.57 &  66.15 &  0.032 & 0.291 \\
        
            Ours-final
                   & \textbf{27.16} & \textbf{63.58}  & \textbf{0.010}  & \textbf{0.287}  \\

        \hline
        \end{tabular}
         \vspace{-0.5em}
        \caption{Quantitative ablation studies on the task of room mask-conditioned scene generation on the 3D-FRONT bedrooms. The top two row is the ablation of the Number of anchor points. The third row ablates the training strategy of our Scene Generation Transformer. } 
        \label{tab:ablation}
        \end{center}  
        \vspace{-3em}
\end{table}

\subsection{Downstream tasks.}
\begin{figure}[!htbp]
    \centering
    \includegraphics[width=0.45\textwidth]{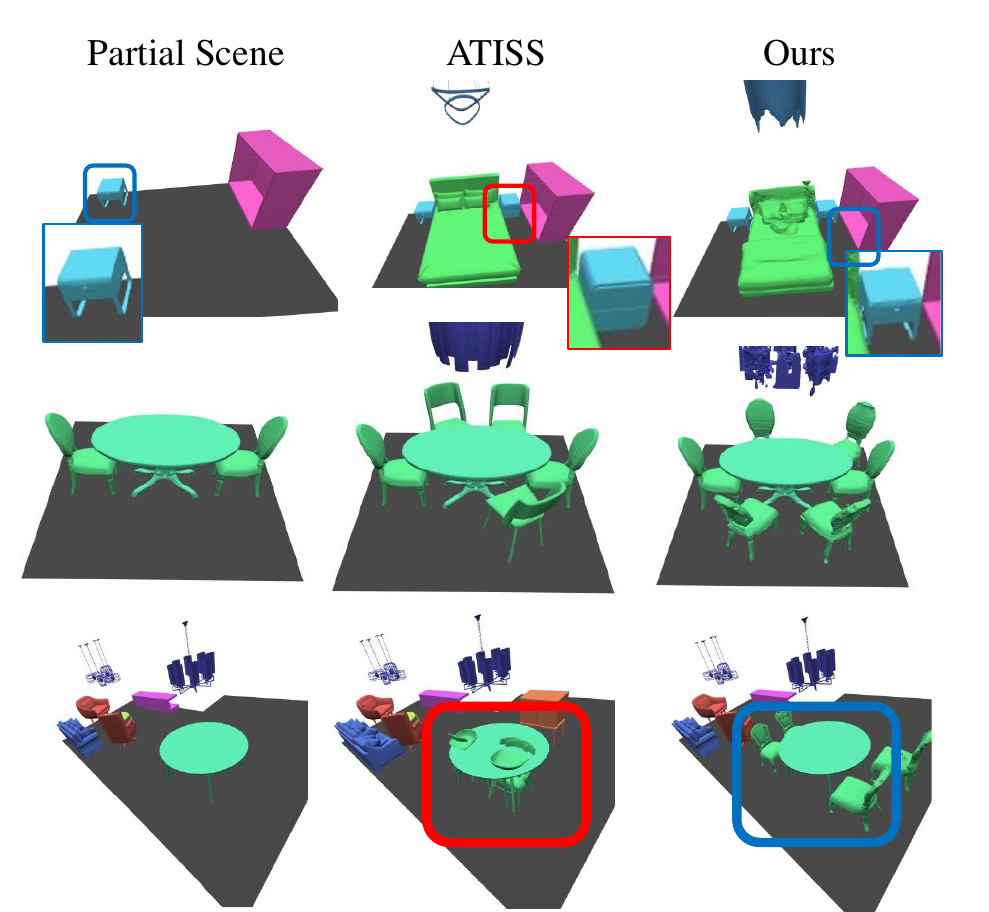}
      \vspace{-3mm} 
    \caption{Style-consistent and Shape-compatible scene completion compared with ATISS~\cite{paschalidou2021atiss}. The blue rectangle means that our completed objects can be style consistent and shape-compatible with given partial objects, while the red rectangle means ATISS can not achieve this.}
     \vspace{-1.5em} 
    \label{fig:completion}
\end{figure}

\noindent\textbf{Style-consistent and shape-compatible scene completion.} As shown in Fig.~\ref{fig:completion}, when given a partial scene with some furniture already in it. We compare our results with ATISS~\cite{paschalidou2021atiss}, and we conduct scene completion based on the room floor plan and existing objects. Since  ATISS~\cite{paschalidou2021atiss} does not take the furniture shape into consideration when given some previous object. ATISS~\cite{paschalidou2021atiss} can not complete the scene with furniture consistent with existing furniture. With our proposed  anchor-latents and shape-aware parameterization, our shape-aware scene generation transformer can complete the scene with a consistent style and compatible furniture. 

\begin{figure}[!tbp]
    \centering
    \includegraphics[width=0.45\textwidth]{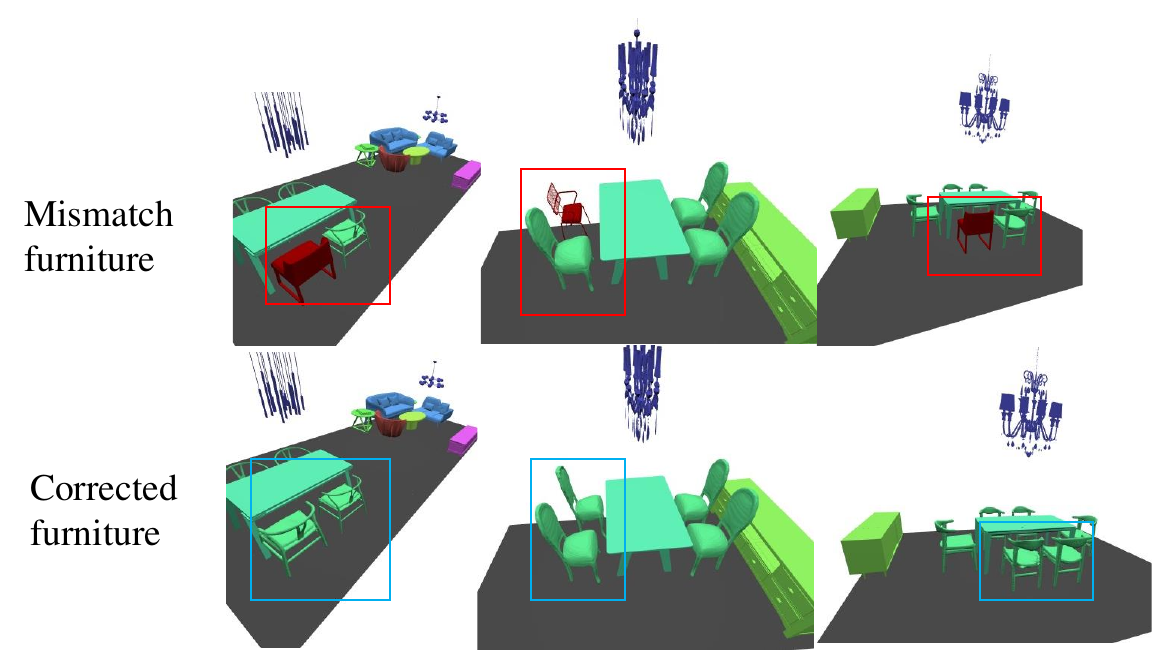}
     \vspace{-3mm} 
    \caption{Mismatch correction. The top row shows the room with mismatched furniture in it. Our method can automatically find the mismatched furniture(in red) and generate style-consistent furniture to correct the shape mismatch.}
     \vspace{-4mm} 
    \label{fig:correctionl}
\end{figure}

\noindent\textbf{Furniture mismatch correction.} As shown in Fig.~\ref{fig:correctionl}, we demonstrate the capability of our model to detect and rectify unnatural inconsistent object styles. For a given scene, we calculate the probability distribution of each object based on our model while considering the presence and shape of other furniture in the scene. In our settings, we designate problematic furniture as those with low likelihood and then proceed to resample a new shape from our generative model in order to rearrange them. Note that this task cannot be
performed by previous methods as they do not consider the shape of each piece of furniture.

\noindent\textbf{Controllable furniture-level editing.} Since our anchor points can control the structure of our furniture, we can interpolate the new shape by mixing the anchor points from two different shapes. Furthermore, our shape-aware scene generation model can further complete the scene with interpolated furniture. Note that this task is all the previous scene synthesis work can not achieve. Please refer to the supplementary materials for the visualization results. 

\noindent\textbf{Discussion and Limitations.} The main limitation of our method is that we do not consider texture. Exploring the combination of state-of-the-art texture generation methods~\cite{chen2023text2tex, richardson2023texture} with our methods may be an interesting future direction for scene generation. 

%% file: sec/5_conclusion.tex
\section{Conclusion}

We introduce Roomdesigner, an indoor scene generation transformer that jointly synthesis the room layout and generates the furniture shape. To achieve this, we propose an anchor-latent representation, a representation that can capture both global structure and local geometry for each piece of furniture. Based on the anchor-latent representation, we learn a scene generation transformer to auto-regressively generate the location information and shape information for each piece of furniture. With this shape-aware scene generation model, we can achieve style-consistent and shape-compatible scene generation. Furthermore, our model can also facilitate a lot of downstream tasks such as shape-compatible and style-consistent scene completion, style-mismatch detection and correction, and controllable furniture-level editing, which can not be achieved by previous works. Experiments verify that our method can achieve compelling results in the room-mask conditional generation task.

%% file: sec/6_suppl.tex
\clearpage

\setcounter{section}{0}
\renewcommand\thesection{\Alph{section}}
\renewcommand\thesubsection{\thesection.\arabic{subsection}}
\renewcommand\thesubsubsection{\thesubsection.\arabic{subsubsection}}
\maketitlesupplementary

In this supplemental material,  we provide details for our implementation in Sec.~\ref{sec:implement}, we provide the results for controllable editing with anchor-latents in Sec.~\ref{sec:edite}, we provide the qualitative results for our ablation studies in Sec.~\ref{sec:ablation}, we provide the implementation details of our baseline in Sec.~\ref{sec:baseline}. Finally, we provide more visual results of our method to prove that our method can enable style-consistent and shape-compatible indoor scene generation results in Sec.~\ref{sec:vis}. We also provide many multi-view visualizations results as\textbf{ $.gif$ in our supplementary files} to show the style-consistent and shape-compatible indoor scene generation results of our methods.

\section{Implementation details.}
\label{sec:implement}
\subsection{Network Architecture and loss functions}
\paragraph{Scene generation transformer.}We follow the design from~\cite{paschalidou2021atiss, yi2022mime} to implement our Scene generation transformer encoder as a multi-head attention transformer without any positional encoding. Our scene generation transformer consists of 4 layers and 16 attention heads. The query, key, and value have 64 dimensions. The total dimension for a single furniture token is, therefore $16\times 64 = 1024$. Where the first 512-dimensions tokens embedded the feature extracted from Bounding Box Encoder, while the next 512-dimensions tokens embedded the feature extracted from Anchor-latents Encoder. The Feed-forward MLP has 2048 dimensions. $q \in \mathbb{R}^{64}$  is a learnable object query vector that predicts the output feature $\hat{q} \in \mathbb{R}^{64}$ used for the next furniture prediction. We implement our transformer using the transformer library~\cite{katharopoulos2020transformers}. The learnable embedding $\theta_{c}$ for the category is implemented with a single layer MLP that maps the one-hot label into $C\in\mathbb{R}^{64}$, while the embedding $\theta_{anchor}$ is implemented with a two-layer MLP that map the anchor embedding from $512\times(4\times64)$ in to $512\times 1$. The loss function for the category is a Cross entropy loss. We follow previous work~\cite{paschalidou2021atiss} and model the size $s\in \mathbb{R}^3$, the translation $t \in \mathbb{R}^3$ and the rotation $r\in \mathbb{R}^1$ with a mixture of logits distributions.
\begin{equation}
    s\sim \sum_{k=1}^K \pi_k^s \text{logistic}(\mu_k^s, \sigma_k^s)
\end{equation}
\begin{equation}
    t\sim \sum_{k=1}^K \pi_k^t \text{logistic}(\mu_k^t, \sigma_k^t)
\end{equation}
\begin{equation}
    r\sim \sum_{k=1}^K \pi_k^r \text{logistic}(\mu_k^r, \sigma_k^r)
\end{equation}

where $\pi_k^s, \mu_k^s, \sigma_k^s$ denote the weight, mean, and variance of the k-th logistic distribution used for modeling the size. Similarly for $t$ and $r$. For implementation, we set $K=10$, following the same design of ATISS~\cite{paschalidou2021atiss}. For the extract of each bounding box attribute, we use a 2-layer MLP to achieve this. We use the negative log-likelihood loss for these attributes. Therefore, the loss function for the layout prediction is 
\begin{equation}
\begin{aligned}
      \mathcal{L}_{\text{layout}} = \text{CE}(c, \hat{c}) + \text{NLL}_{\text{size}}(s, \hat{s}) + \\ \text{NLL}_{\text{translation}}(t, \hat{t}) +  \text{NLL}_{\text{rotation}}(r, \hat{r})
\end{aligned}
\end{equation}

\paragraph{Shape Generation Transformer.}
\begin{figure}[h!]
    \centering
    \includegraphics[width=0.45\textwidth]{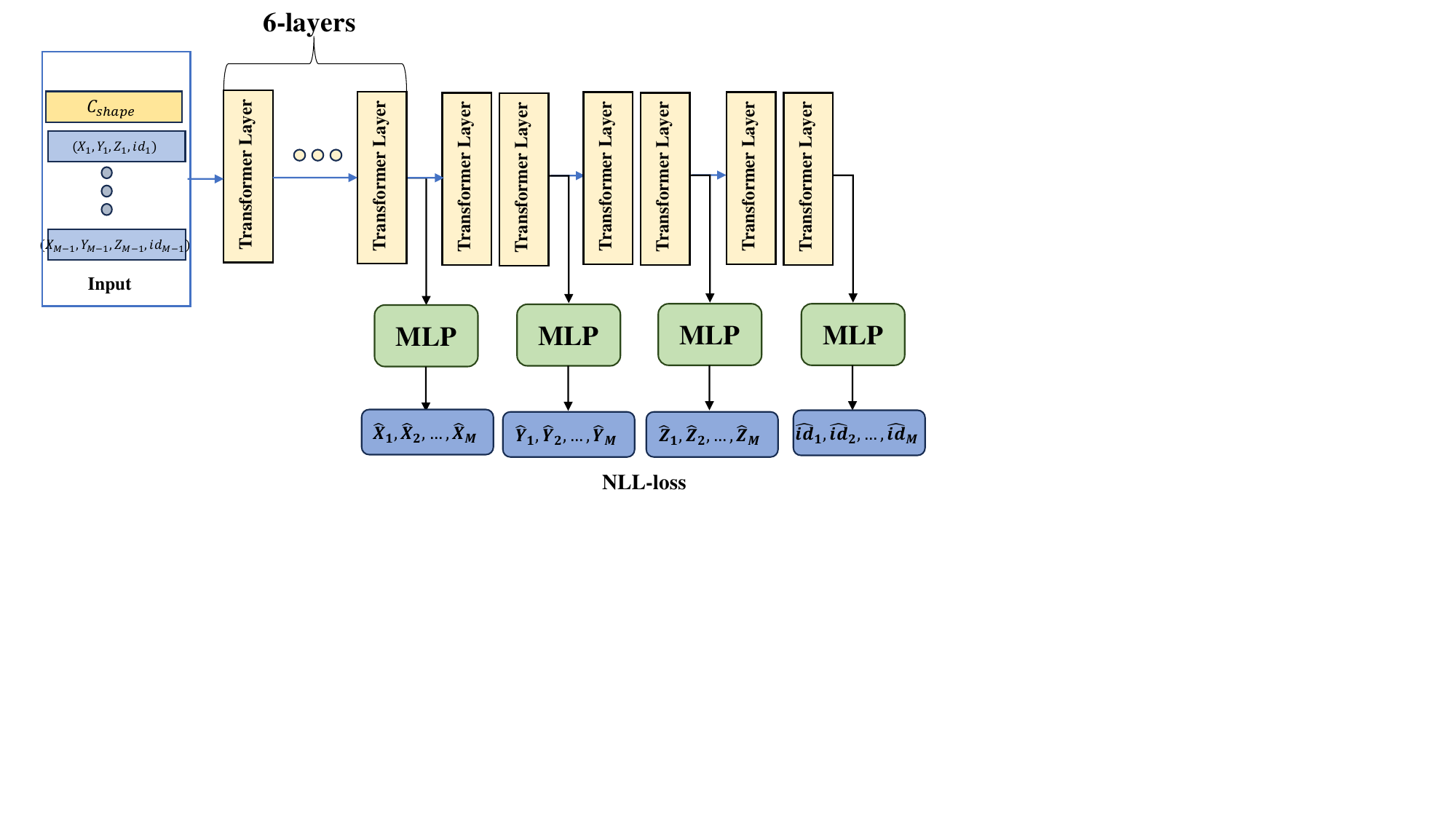}
    \caption{Network details for shape generation transformer}
    \label{fig:shape_trans}
\end{figure}

The shape generation transformer is implemented with 12 layers in total, as shown in Fig.~\ref{fig:shape_trans}. The length for each token is 512, which corresponds to the number of our anchor-latents $M$. For the training stage,  We first embedded the coordinate $x, y, z$ for each anchor point and the quantized index $id$ for each anchor feature in the codebook, each as $M\times 512$ using the learnable embeddings, where the second 512 means the feature dimension for each learnable embeddings. 
We collect four features, each of size $M\times 512$ feature for $x, y, z, id$, by summing them together. We then concatenate $\mathcal{C}_{shape}\in \mathbb{R}^{512}$ with the first $M-1$ feature for the previous $M-1$ anchor-latents to get the input embedding. Each embedding is used to predict the next anchor-latents. We follow the same design of the bounding box extractor that predicts the $x, y, z, id$ sequentially. Specifically, the first six layers was used to predict the embedding for $x$, the next two layer for $y$, then for $z$ and $id$ similarly after each feature prediction. We decode them using a separate two-layer MLP. Similar to the attribute extractor, we model the distribution of the anchor points $x, y, z$ using a mixture of logit distributions.
\begin{equation}
    x\sim \sum_{k=1}^K \pi_k^x \text{logistic}(\mu_k^x, \sigma_k^x)
\end{equation}
\begin{equation}
    y\sim \sum_{k=1}^K \pi_k^y \text{logistic}(\mu_k^y, \sigma_k^y)
\end{equation}
\begin{equation}
    z\sim \sum_{k=1}^K \pi_k^z \text{logistic}(\mu_k^z, \sigma_k^z)
\end{equation}
Therefore, for the shape generation branch optimization, we use the negative log-likelihood loss for the quantized id classification and use the negative log-likelihood loss for the distribution for anchor points coordinate.
\begin{equation}
\begin{aligned}
      \mathcal{L}_{\text{shape}} = \text{NLL}(id, \hat{id}) + \text{NLL}_{\text{x}}(x, \hat{x}) + \\ \text{NLL}_{\text{y}}(y, \hat{y}) +  \text{NLL}_{\text{z}}(z, \hat{z})
\end{aligned}
\end{equation}

During inference, starting from $\mathcal{C}_{shape}$, we autoregressively predict the next anchor-latent from 0 until  $M=512$.

\subsection{Training Details}
For the shape VQVAE training, we train our model on 4 NVIDIA A40 GPUs with a batch size of 16 on each GPU. We train our VQVAE for 1000 epochs with Adam Optimizer. The learning rate for training is 1e-3. The number of sample points $N=2048$, the number of latent anchors $M=512$, $k=32$ for $k$-NN The size of the codebook is $|\mathcal{D}| = 1024$, where the dimension for each latent is $C=256$.

For the scene generation stage, we train our model with a batch size of 128 for 500k iterations. We use the optimizer, Adam. The learning rate for training is 1e-3. The training process is with no weight decay. The scene generation stage was trained on 8 NVIDIA A40 GPUs. 



\subsection{Dataset processing details}

\paragraph{Furniture VQ-VAE stage.}To train the occupancy fields, the mesh should be watertight. We first use O-CNN~\cite{wang2017ocnn} to extract the watertight surface of each Furniture in the 3D-Future~\cite{fu20213dfuture} dataset. All the vertices of mesh was scaled into $[-0.95, 0.95]$. For optimization of the first stage, we randomly select points in the volume and use pysdf\footnote{https://github.com/andreasBihlmaier/pysdf} to get the ground truth occupancy for each point.

\paragraph{Scene Generation stage.}The dataset pre-processing is based on the setting of ATISS~\cite{paschalidou2021atiss}; we initiate the process by excluding scenes exhibiting intricate object configurations, notably marked by erroneous object class attributions. For instance, instances, where beds are inaccurately categorized as wardrobes in certain scenes are rectified. Following this, scenes manifesting unnatural scales are subsequently eliminated from consideration. Consequently, we disregard scenes characterized by an inadequate or excessive object count. For valid bedroom configurations, the prescribed object range spans from 3 to 13. In the case of dining and living rooms, the lower and upper bounds for object quantities are established at 3 and 21, respectively. Furthermore, scenes containing objects falling outside the confines of predetermined categories are expunged. Following the pre-processing phase, our dataset comprises 4,041 bedrooms, 900 dining rooms, and 813 living rooms. We have additional classes of  `begin' and 'end' to define the start token and the end token of our autoregressive prediction. Combining with the object categories that appeared in each room type, we have $L=23$ object categories for bedrooms and $L=26$ object categories for dining and living rooms in total. The category labels are listed as follows.
\\
\begin{python}
# 23 3D-Front bedroom categories
[ 'armchair', 'bookshelf', 'cabinet',
'ceiling_lamp', 'chair', 'children_cabinet',
'coffee_table', 'desk', 'double_bed',
'dressing_chair', 'dressing_table', 'kids_bed',
'nightstand', 'pendant_lamp', 'shelf',
'single_bed', 'sofa', 'stool', 'table',
'tv_stand', 'wardrobe', 'start', 'end' ]

# 26 3D-Front dining or living room categories
['armchair', 'bookshelf', 'cabinet', 
'ceiling_lamp', 'chaise_longue_sofa', 
'chinese_chair', 'coffee_table', 'console_table',  
'corner_side_table',  'desk', 'dining_chair', 
'dining_table', 'l_shaped_sofa', 'lazy_sofa', 
'lounge_chair', 'loveseat_sofa', 
'multi_seat_sofa', 'pendant_lamp', 
'round_end_table', 'shelf', 'stool', 
'tv_stand', 'wardrobe', 'wine_cabinet', 
'start', 'end' ]
\end{python}

\subsection{Evaluation details}
\paragraph{Collision Rate Calculation.}We randomly generate 1000 scenes using the room mask from the test set. We denote $N_{i}$ as the number of the generated furniture in the $i-th$ generated room. We denote $C_{i}$ as the number of the furniture which collided with other furniture in the $i-th$ generated room. The Collision Rate is calculated with the following:
\begin{equation}
    coll = \frac{1}{1000}\sum_{i=1}^{1000} \frac{C_i}{N_i}
\end{equation}

\paragraph{Shape-consistency evaluation.}We randomly generate 1000 scenes using the room mask from the test set. For the specific furniture category, we calculate the average Chamfer-Distance and Open-Shape similarity score among the furniture that shares the same category in a single generated scene. Then we calculated the average score across 1000 generated scenes.

\paragraph{Shape-diversity evaluation.} For each room mask in the test set,  we generate ten different scenes conditional on that room mask. Then for each selected furniture category(e.g., Bed and Nightstand for bedroom, Chair and Table for dining room and livingroom). We randomly selected a single piece of furniture within the predefined category for each generated scene. Consequently, for each room mask, the average Chamfer-Distance is computed for this selected furniture item. Subsequently, the mean Chamfer Distance is calculated across distinct room masks.

\section{Controllable furniture-level to scene-level editing with Anchor-latents.}
\label{sec:edite}

\begin{figure*}
    \centering
    \includegraphics[width=0.9\textwidth]{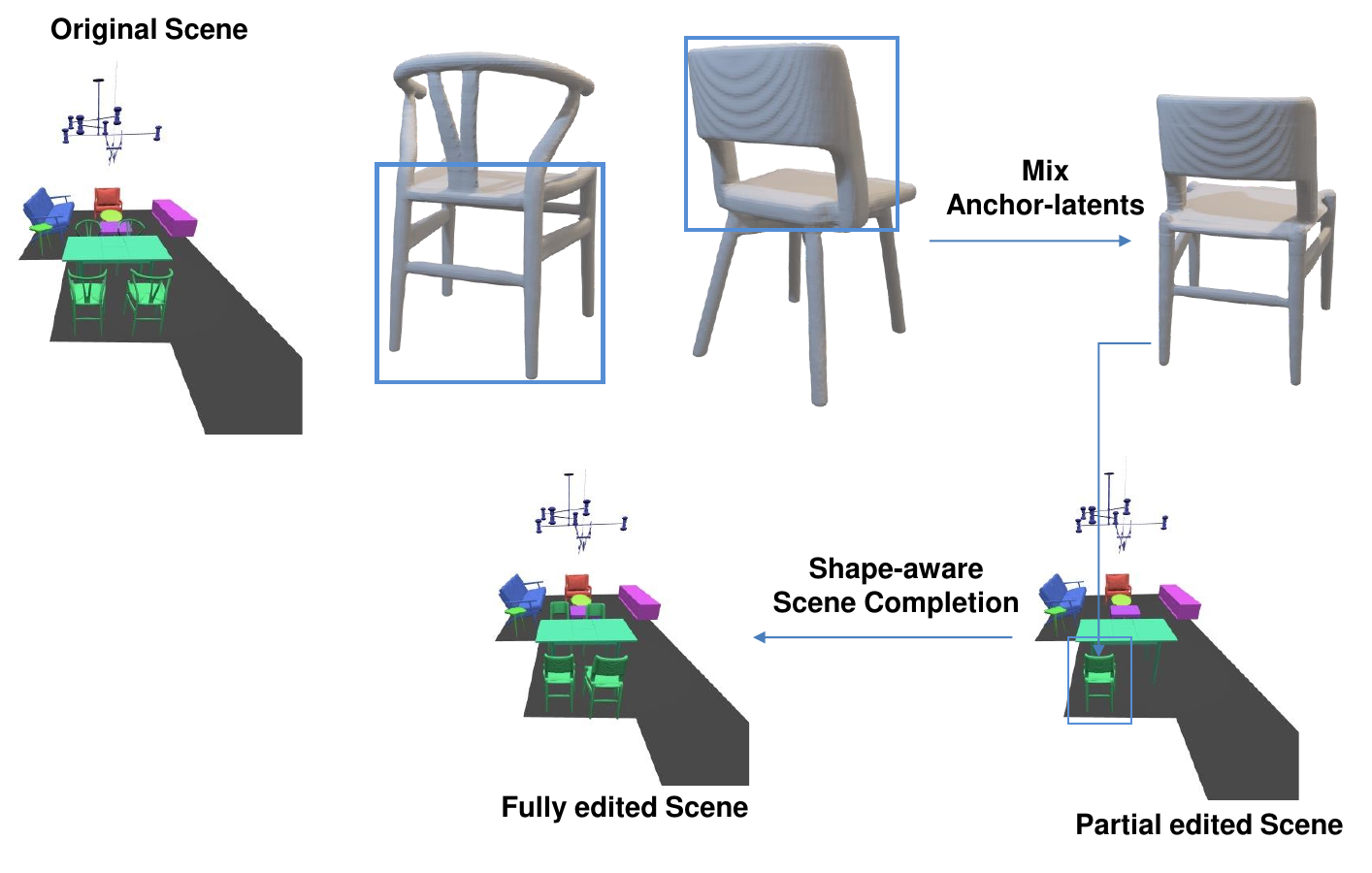}
    \caption{Furniture-level scene editing results: By mixing the part anchor-latents from two different furniture items, we can interpolate to the novel furniture shape, then we can conduct shape-aware scene completion to achieve scene editing. }
    \label{fig:edit1}
\end{figure*}
\begin{figure*}
    \centering
    \includegraphics[width=0.9\textwidth]{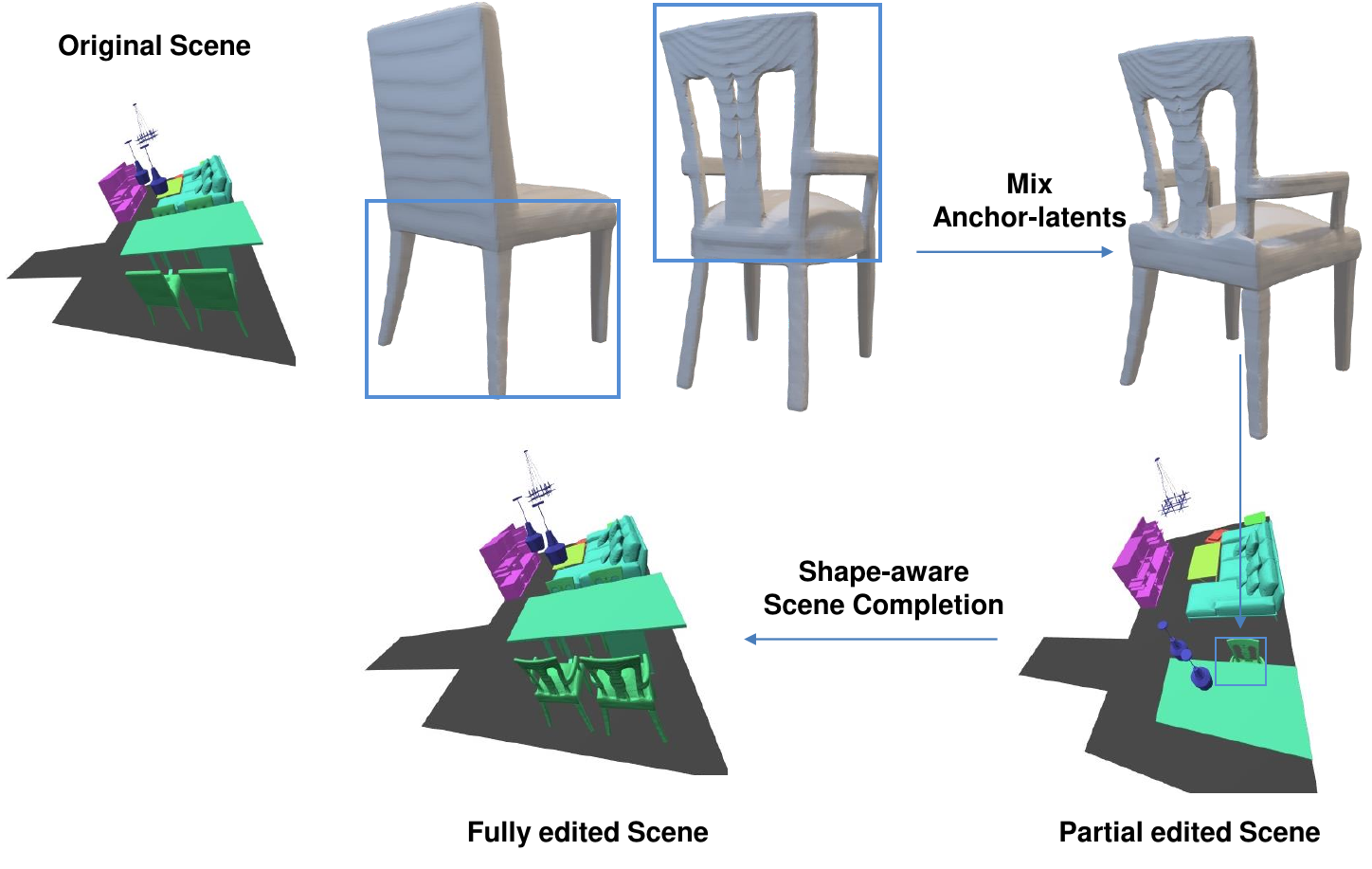}
    \caption{Furniture-level scene editing results: By mixing the part anchor-latents from two different furniture items, we can interpolate to the novel furniture shape, then we can conduct shape-aware scene completion to achieve scene editing. }
    \label{fig:edit2}
\end{figure*}
\begin{figure*}
    \centering
    \includegraphics[width=0.9\textwidth]{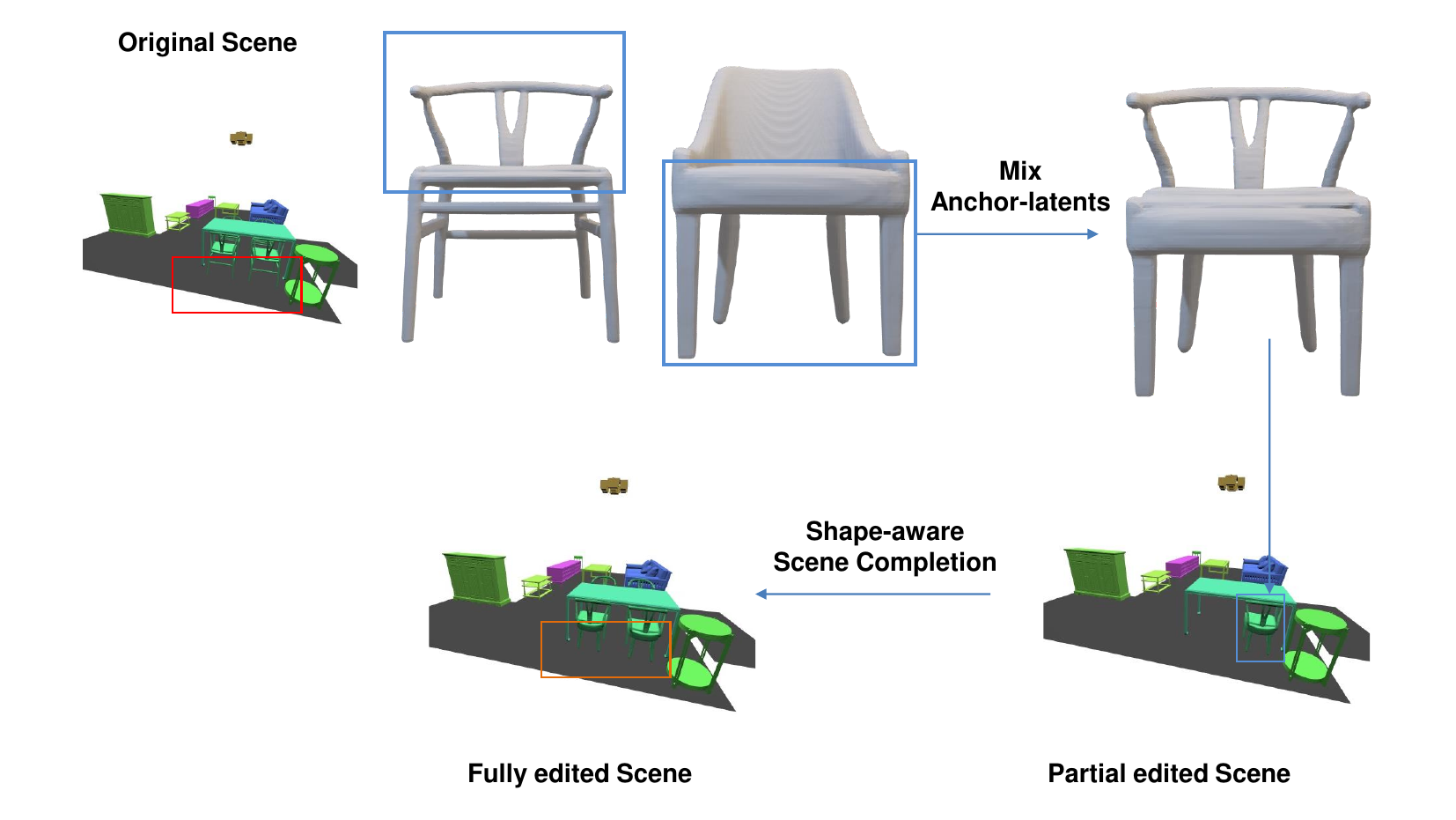}
    \caption{Furniture-level scene editing results: By mixing the part anchor-latents from two different furniture items, we can interpolate to the novel furniture shape, then we can conduct shape-aware scene completion to achieve scene editing. }
    \label{fig:edit3}
\end{figure*}
\begin{figure*}
    \centering
    \includegraphics[width=0.9\textwidth]{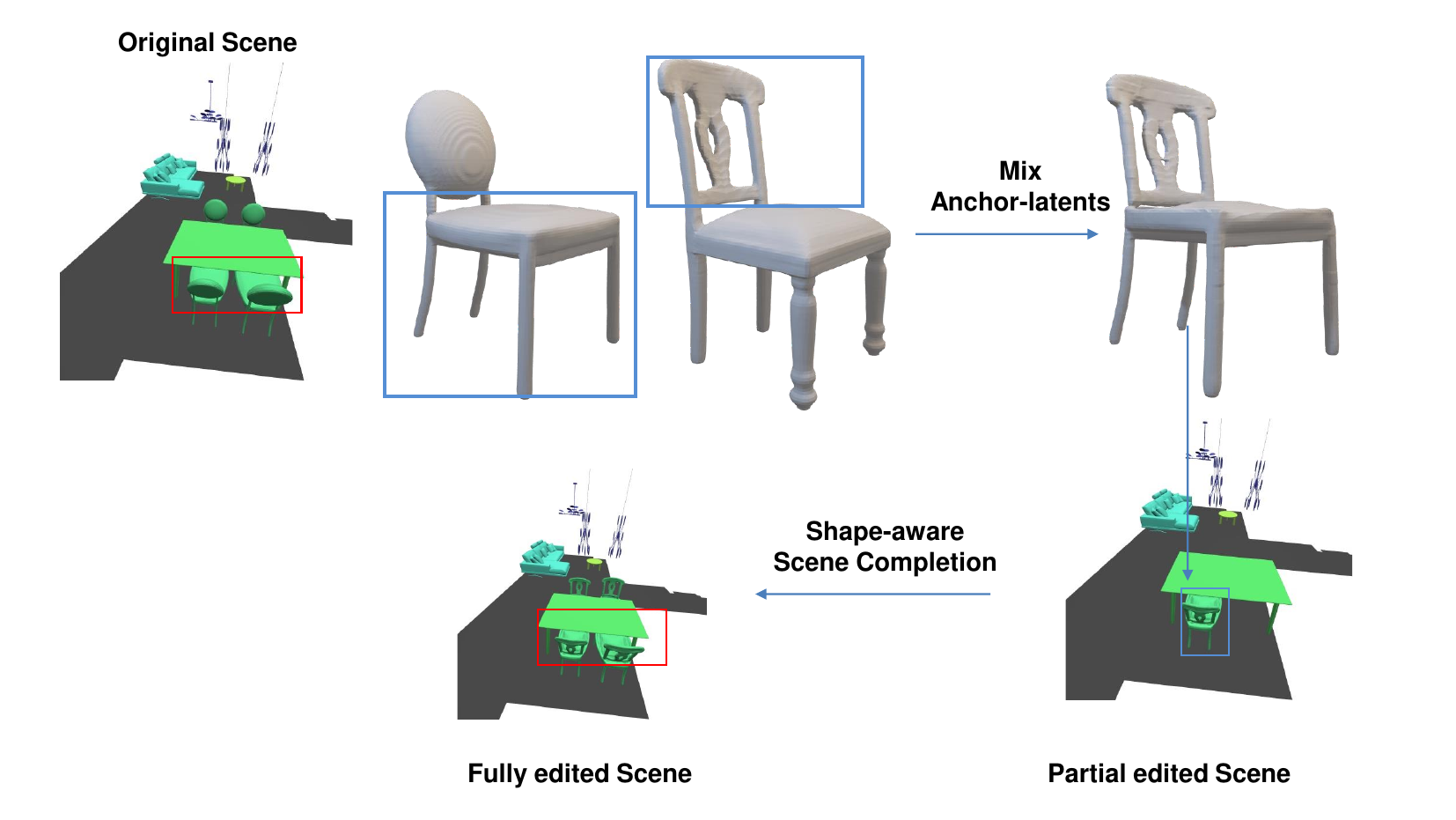}
    \caption{Furniture-level scene editing results: By mixing the part anchor-latents from two different furniture items, we can interpolate to the novel furniture shape, then we can conduct shape-aware scene completion to achieve scene editing. }
    \label{fig:edit4}
\end{figure*}
Since our anchor points can control the structure of our furniture, we can interpolate the new shape by mixing the anchor-latents from two different shapes. Furthermore, our shape-aware scene generation model can further complete the scene with interpolated furniture. Specifically, we randomly select scenes with repetitive furniture in them from the test set. Then, we choose another piece of furniture from the 3D-Future~\cite{fu20213dfuture} dataset. We encode both pieces of furniture as ordered anchor-latents. We then replace the partial anchor-latents of the existing furniture with the partial anchor-latents from another piece of furniture. We then decode them into a novel piece of furniture, which is a mixture of both pieces of furniture. Then we replace one of the original pieces of furniture with our interpolated novel furniture and remove all the other repetitive furniture. Then, we conduct scene completion on the partial scene. We can get the edited scene results as shown in Fig.~\ref{fig:edit1}, Fig.~\ref{fig:edit2}, Fig.~\ref{fig:edit3}, Fig.~\ref{fig:edit4}.  Note that this task is all the previous scene synthesis work can not achieve. We can achieve interpolated shape generation due to our proposed anchor-latents. We can enable scene-level editing using scene completion due to our proposed shape-aware scene generation transformer. Please also see the video for comparison before($*video\_ori.gif$) and after ($*video\_edit.gif$) scene editing.

\section{Qualitative ablation studies}
\label{sec:ablation}

\begin{figure*}[!ht]
    \centering
    \includegraphics[width=\textwidth]{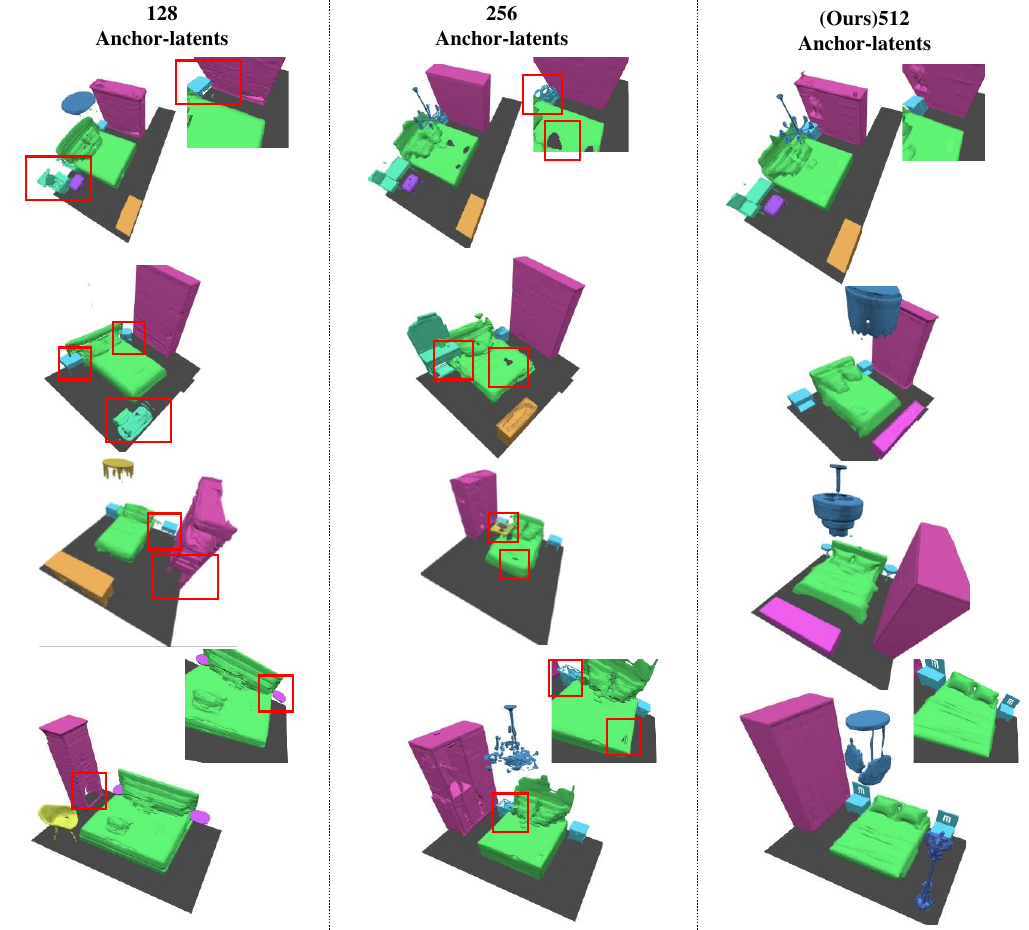}
    \caption{The Qualitative results on 3D-Front Bedroom for ablation study on the number of anchor points. From left to right is the anchor points number 128, 256, and 512(ours) conditioned on the same room mask. Red rectangles indicate that fewer anchor-latents may lead to incomplete shape generation(e.g., no legs or holes in the bed), unreasonable layout prediction, and inconsistent shape generation.}
    \label{fig:ablation}
\end{figure*}

We show the qualitative ablation studies mentioned in the paper. As shown in Fig.~\ref{fig:ablation},  We have found that using fewer anchor-latent easily leads to incomplete shape generation. This is because our sampling is based on point cloud density. If there are not enough anchor points, the anchor-latents used to control finer structures become limited, thus resulting in incomplete shape generation. On the other hand, fewer anchor-latents limit the ability to control the structure of each piece of furniture, thus resulting in inconsistent furniture generation. Furthermore, we have also observed that reducing the number of anchor-latents does not significantly improve layout optimization. This is because fewer anchor-latents do not enable our scene-generation transformer to learn the geometric relationships between furniture items within the scene.

\section{Baseline details}
\label{sec:baseline}
\paragraph{Syn2Gen} Sync2Gen~\cite{yang2021scene} encodes scene arrangements into sequences of 3D objects distinguished by diverse attributes, such as bounding boxes and class categories. Central to their approach is a variational auto-encoder network through which they acquire knowledge of the relative attributes of objects. Additionally, a subsequent phase of Bayesian optimization is employed for post-processing, facilitating the enhancement of object arrangements guided by the priors learned from relative attributes. Since Sync2Gen~\cite{yang2021scene} are designed for unconditional synthesis, for fair comparison, we replace the VAE as conditional VAE~\cite{NIPS2015_8d55a249} conditioned on the room mask. We use the ResNet-18 as a room mask encoder same as ours.

\paragraph{ATISS} ATISS~\cite{paschalidou2021atiss} treats a scene as an unstructured collection of objects and introduces an innovative autoregressive transformer framework to capture the scene generation procedure. In the training phase, leveraging existing object attributes, ATISS employs a permutation-invariant transformer mechanism to combine their characteristics. This enables the model to anticipate the potential attributes of the subsequent object, including its position, dimensions, orientation, and categorical classification, all conditioned on the amalgamated feature representation. We evaluate it using its original settings. Different from our methods, ATISS~\cite{paschalidou2021atiss} only predicts the bounding box information for each generated scene and retrieves the most relevant furniture with the closet bounding box size. Our method directly generates each piece of furniture.

\paragraph{ATISS+OpenShape} We consider injecting shape embeddings into ATISS to make its generation process more shape-aware. OpenShape~\cite{liu2023openshape} is a state-of-the-art method designed for most similar shape retrieval. Its embeddings were pre-trained on different large-scale 3D dataset~\cite{shapenet2015, fu20213dfuture, Deitke_2023_CVPR} and had strong generalization ability. To implement this, We concatenate the original bounding box information of ATISS~\cite{paschalidou2021atiss} with the Openshape~\cite{liu2023openshape} pre-trained embedding. For furniture retrieval, we perform the nearest neighbor search in the 3D-FUTURE~\cite{fu20213d} furniture with the same class label, the closest bounding box size, and the openshape embeddings.

\section{Additional Visualization results}
\label{sec:vis}

\begin{figure*}[h!]
    \centering
    \includegraphics[width=0.9\textwidth]{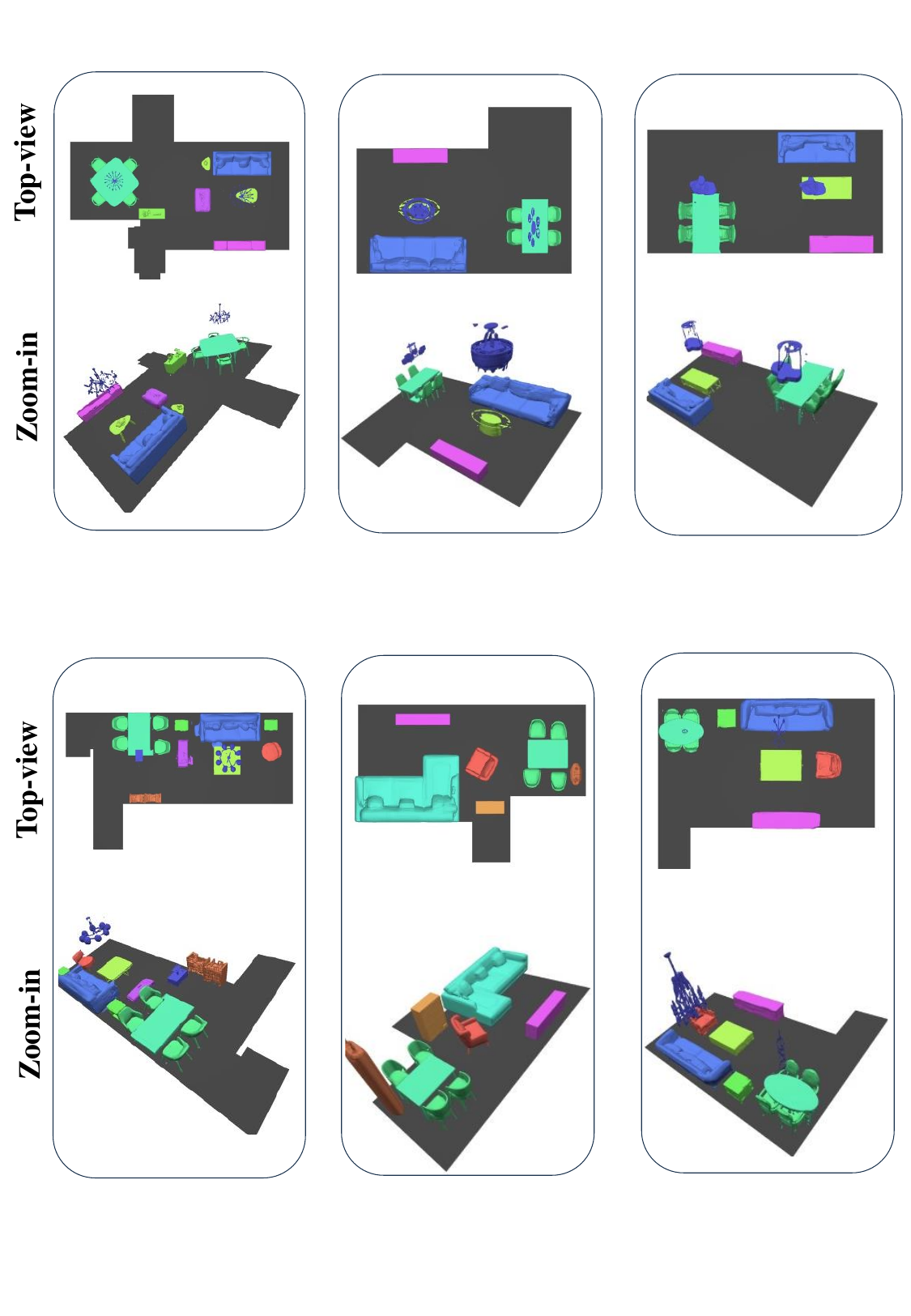}
    \vspace{-10mm}
    \caption{The Qualitative results on 3D-Front Livingroom(1).}
    \label{fig:living1}
\end{figure*}

\begin{figure*}[h!]
    \centering
    \includegraphics[width=0.9\textwidth]{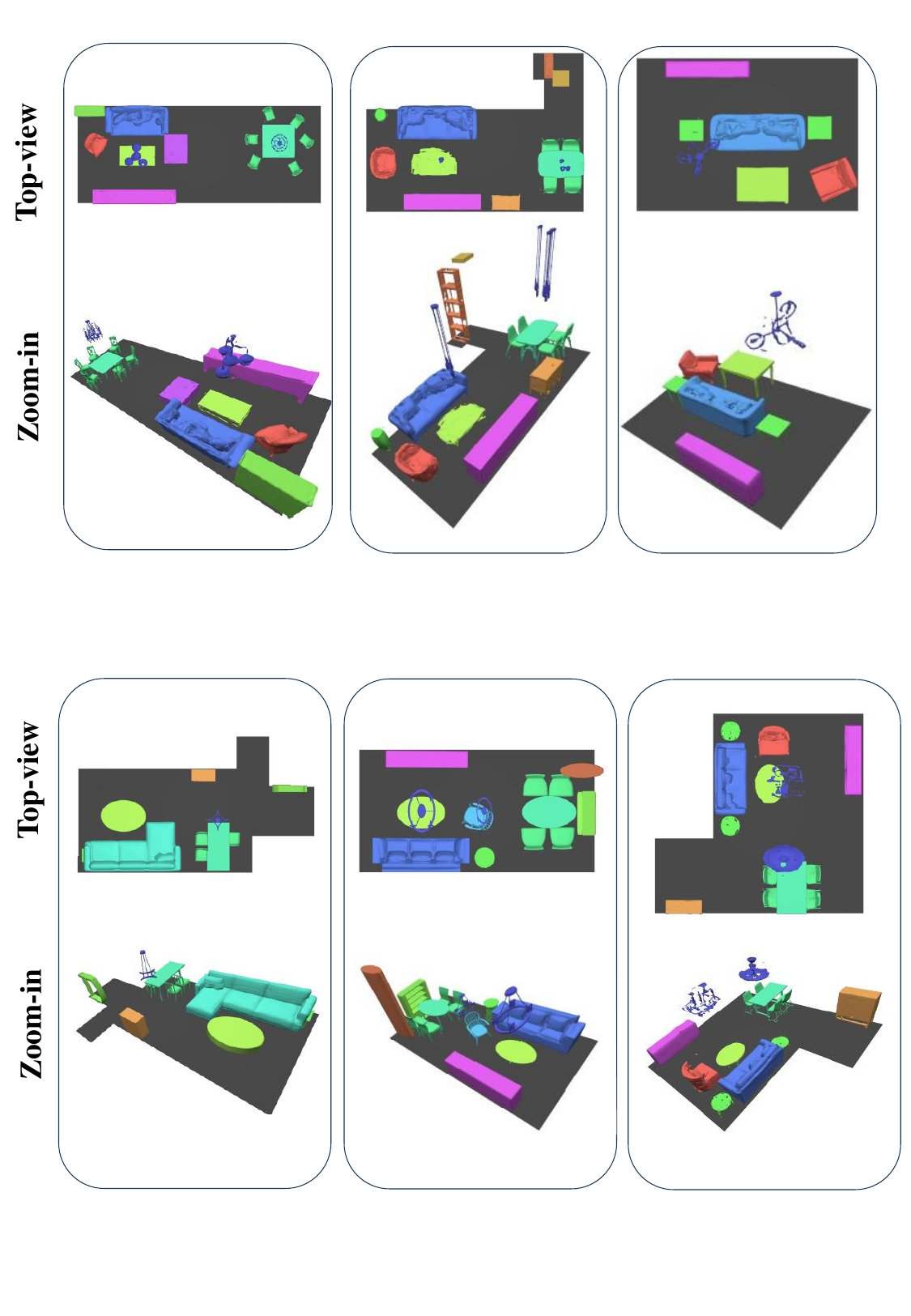}
    \vspace{-10mm}
    \caption{The Qualitative results on 3D-Front Livingroom(2).}
    \label{fig:living2}
\end{figure*}

\begin{figure*}[h!]
    \centering
    \includegraphics[width=0.9\textwidth]{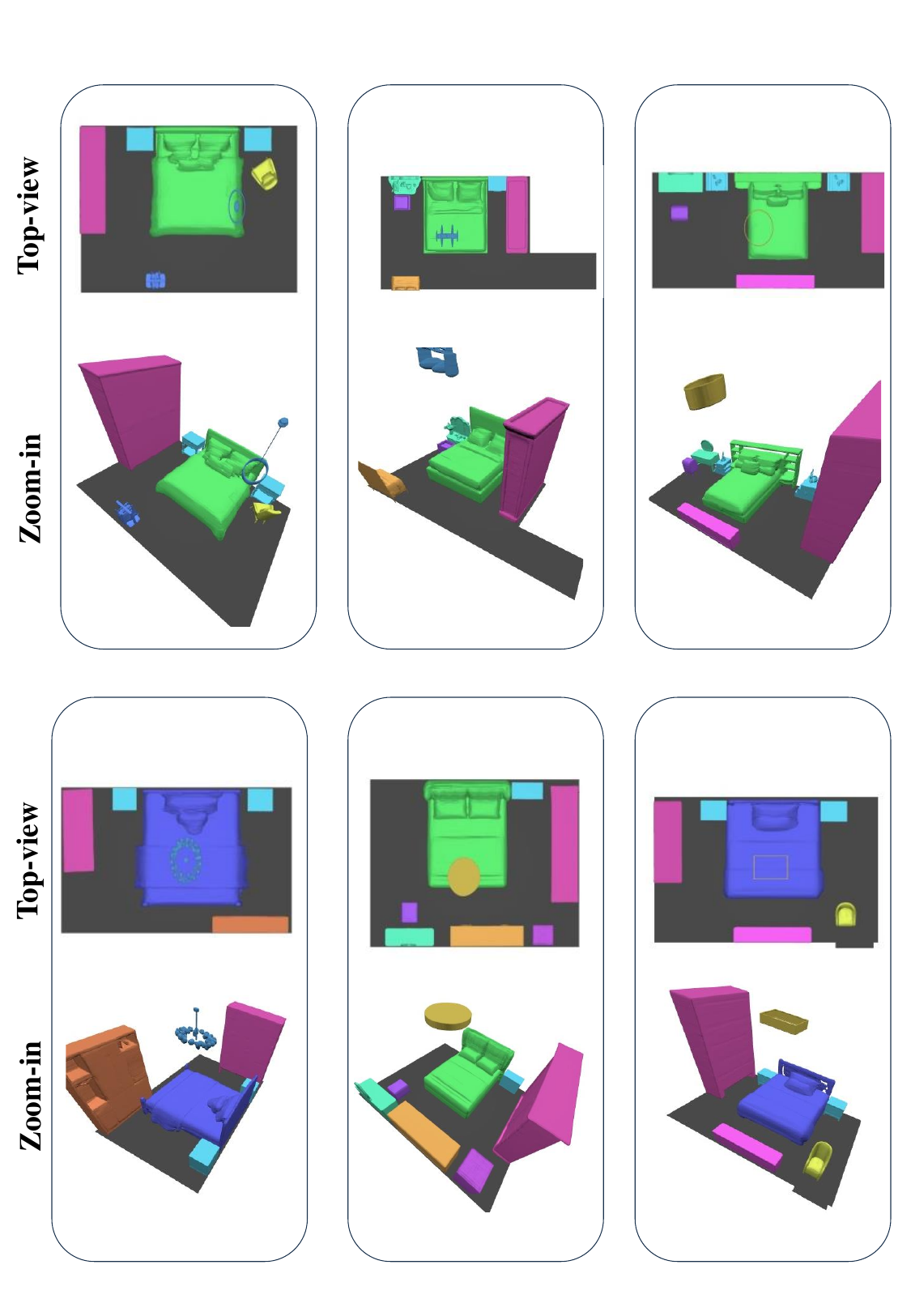}
    \vspace{-10mm}
    \caption{The Qualitative results on 3D-Front Bedrooms(1).}
    \label{fig:bedroom1}
\end{figure*}

\begin{figure*}[h!]
    \centering
    \includegraphics[width=0.9\textwidth]{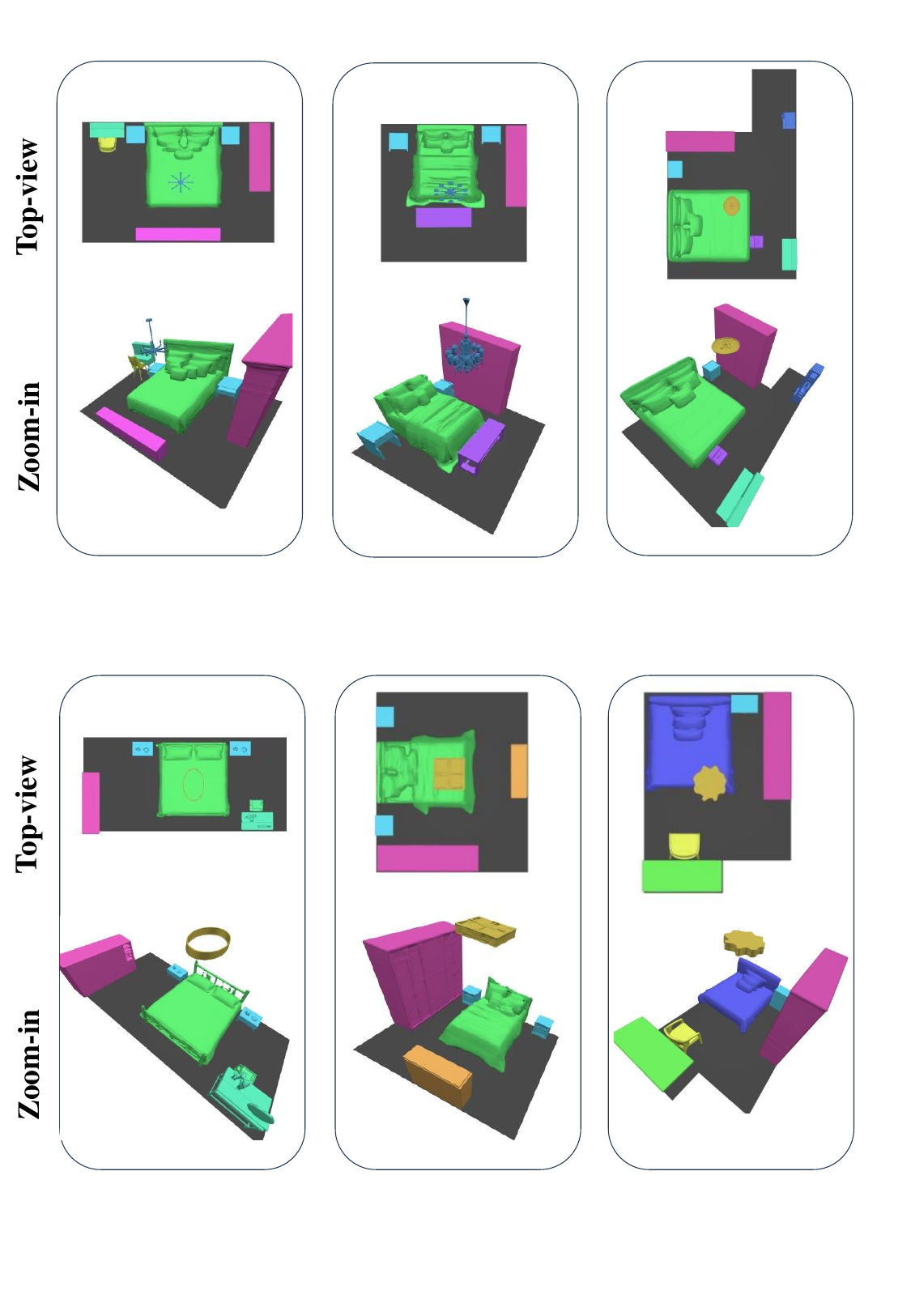}
    \vspace{-10mm}
    \caption{The Qualitative results on 3D-Front Bedrooms(2).}
    \label{fig:bedroom2}
\end{figure*}

\begin{figure*}[h!]
    \centering
    \includegraphics[width=0.9\textwidth]{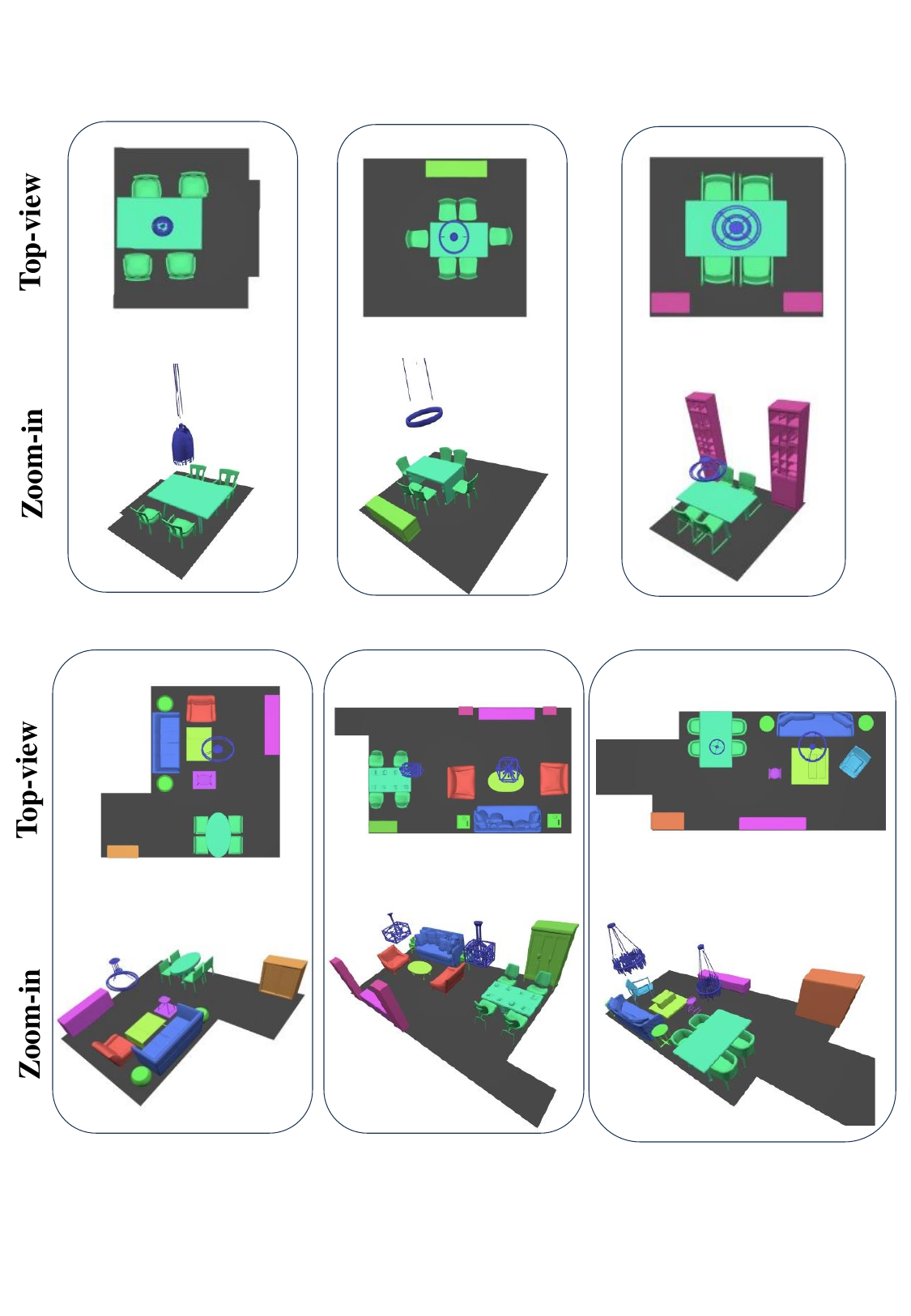}
    \vspace{-10mm}
    \caption{The Qualitative results on 3D-Front Diningroom(1).}
    \label{fig:dining1}
\end{figure*}

\begin{figure*}[h!]
    \centering
    \includegraphics[width=0.9\textwidth]{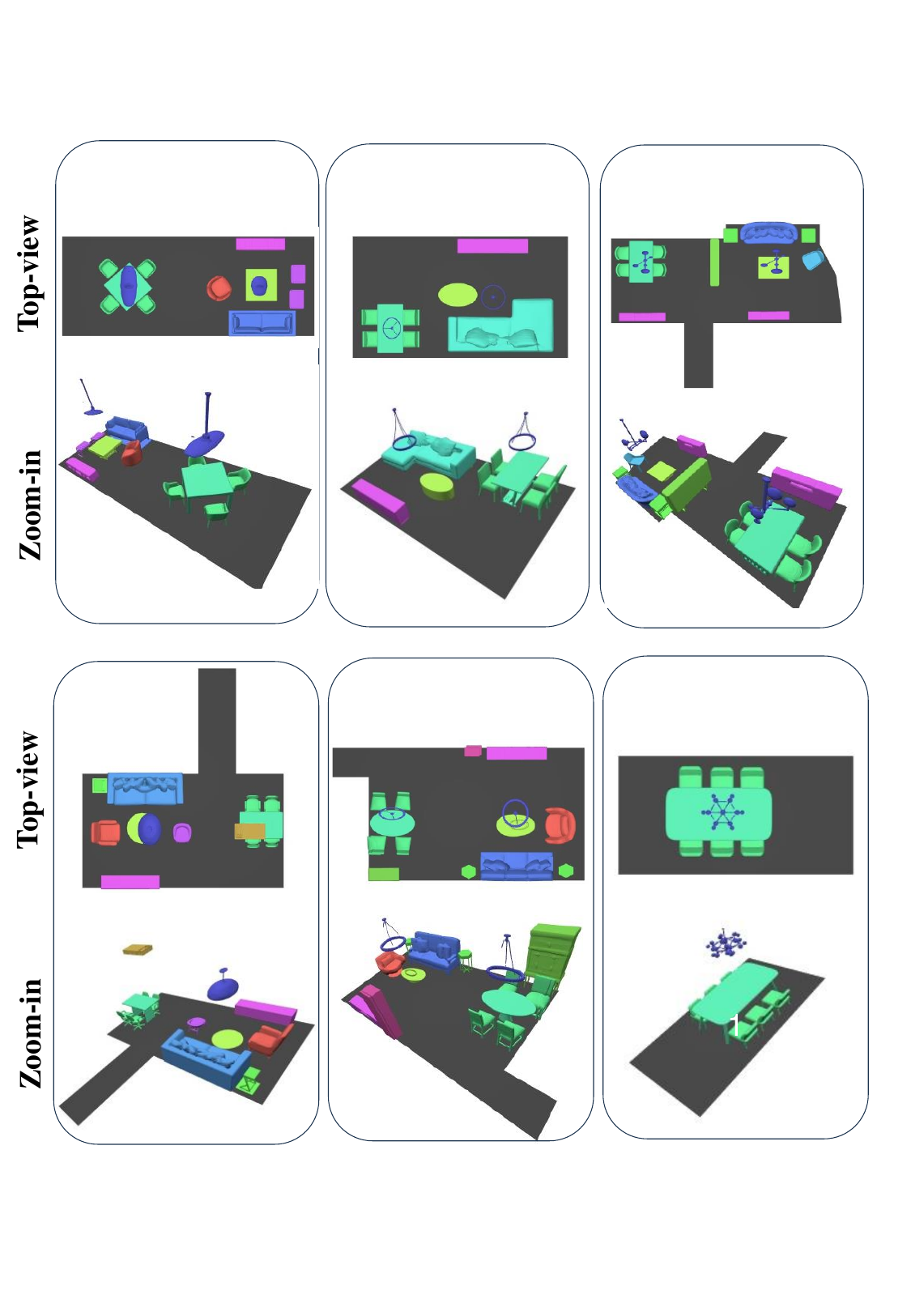}
    \vspace{-10mm}
    \caption{The Qualitative results on 3D-Front Diningroom(2).}
    \label{fig:dining2}
\end{figure*}

We present the results of our room mask-conditioned scene generation across various room types, Livingrooms(Fig.~\ref{fig:living1}, Fig~\ref{fig:living2}), Bedrooms(Fig.~\ref{fig:bedroom1}, Fig~\ref{fig:bedroom2}),  Diningrooms(Fig.~\ref{fig:dining1}, Fig~\ref{fig:dining2}). These results serve to substantiate the high quality of our generated results, concurrently upholding the principles of style-consistent and shape-compatible scene generation. We also provide the multi-view results of our generated scene, please see the $.gif$ files for multi-view visualization results. We can see from these results that our method can also improve the scene diversity. even if the layour is limited(bedroom), our method can also generate diverse shape for each piece of furniture.

\clearpage